\documentclass{article}

\usepackage{arxiv}
\usepackage[square,numbers,sort,compress]{natbib}
\usepackage[utf8]{inputenc} 
\usepackage[T1]{fontenc}    
\usepackage{hyperref}       
\usepackage{url}            
\usepackage{booktabs}       
\usepackage{amsfonts}       
\usepackage{nicefrac}       
\usepackage{microtype}      
\usepackage{lipsum}		
\usepackage{graphicx}
\usepackage{doi}
\usepackage{tabularx}
\usepackage{graphicx}
\usepackage{subcaption}
\usepackage{amsmath}
\usepackage{mdframed}
\usepackage{longtable}
\newcolumntype{P}[1]{>{\raggedright\arraybackslash}p{#1}}
\usepackage{comment}
\usepackage{csquotes}
\usepackage[table]{xcolor}
\usepackage{multirow}
\usepackage{array}
\usepackage{listings}
\usepackage{soul}
\usepackage{wrapfig}
\usepackage{anyfontsize}
\usepackage{orcidlink}

\usepackage{soul}
\usepackage{xcolor}

\usepackage{fancyvrb}

\title{Image First or Text First? Optimising the Sequencing of Modalities in Large Language Model Prompting and Reasoning Tasks}

\author{ 
\orcidlink{https://orcid.org/0009-0009-6026-2321} Grant Wardle \\
	School of Mathematical and Computational Sciences\\
	Massey University (Alumni)\\
	Albany, New Zealand \\  
	\And
   \orcidlink{https://orcid.org/0000-0001-9416-1435} Teo ~Susnjak\thanks{Corresponding author: t.susnjak@massey.ac.nz} \\
	School of Mathematical and Computational Sciences\\
	Massey University\\
	Albany, New Zealand \\ 
}

\renewcommand{\shorttitle}{Sequencing of Modalities in Large Language Model Prompting}

\begin{document}
\maketitle

\begin{abstract}

This paper examines how the sequencing of images and text within multi-modal prompts influences the reasoning performance of large language models (LLMs). We performed empirical evaluations using three commercial LLMs. Our results demonstrate that the order in which modalities are presented can significantly affect performance, particularly in tasks of varying complexity. For simpler tasks involving a single image, modality sequencing had a clear impact on accuracy. However, in more complex tasks involving multiple images and intricate reasoning steps, the effect of sequencing diminished, likely due to the increased cognitive demands of the task. 
Our findings also highlight the importance of question/prompt structure. In nested and multi-step reasoning tasks, modality sequencing played a key role in shaping model performance. While LLMs excelled in the initial stages of reasoning, they struggled to re-incorporate earlier information, underscoring the challenges of multi-hop reasoning within transformer architectures. This suggests that aligning the sequence of modalities with the logical flow of reasoning steps is more critical than modality order alone. These insights offer valuable implications for improving multi-modal prompt design, with broader applications across fields such as education, medical imaging, and cross-modal learning.


\end{abstract}


\section{Introduction}

Recent advancements in Large Language Models (LLMs) have profoundly impacted natural language understanding and related fields seeking to automate tasks involving human language. While reasoning was once considered a uniquely human trait \cite{royce2019teaching}, parallels are now observed between human cognition and LLMs. The emergent reasoning abilities of LLMs and solve complex tasks that require high-order cognitive abilities has generated significant academic attention \cite{NEURIPS2023_117c5c86,chen-etal-2024-m3cot}, as well as concerns in some fields \cite{McIntosh2024game} about the trajectory of such AI agents. Considerable research efforts have been devoted to improving LLMs' reasoning abilities recently which, while impressive, have nevertheless been uneven and variable across different tasks \cite{NEURIPS2023_117c5c86}. With the emergence of multi-modal LLMs that can now process textual, audio and visual inputs, the complexity of reasoning across all modalities has increased markedly \cite{feng2024more,pal2024gemini,stribling2024model} and questions remain about how best to structure the prompts to elicit optimal reasoning in such contexts. 

Visual question-answering (VQA) involving a combination of image(s) and multiple-choice questions has become a common method for evaluating LLM multi-modal reasoning capabilities \cite{liu2024mmbenchmultimodalmodelallaround, lu2022learnexplainmultimodalreasoning, chen-etal-2024-m3cot, zhang2023m3exammultilingualmultimodalmultilevel, li2023seedbenchbenchmarkingmultimodalllms}. Benchmark datasets for this task have emerged \cite{chen-etal-2024-m3cot} covering a wide range of disciplines while often taking the form of academic exam-like questions \cite{zhang2023m3exammultilingualmultimodalmultilevel}. Notably, models such as GPT-4 \cite{openai2024gpt4technicalreport}, Gemini-1.5 \cite{reid2024gemini}, and Claude \cite{anthropic2024claude} have displayed degrees of multi-modal reasoning in the context of VQA \cite{liu2024mmbenchmultimodalmodelallaround, lu2022learnexplainmultimodalreasoning, chen-etal-2024-m3cot, zhang2023m3exammultilingualmultimodalmultilevel, li2023seedbenchbenchmarkingmultimodalllms}; however, these capabilities are not well understood. Initial research has indicated that LLMs significantly struggle with multi-modal reasoning tasks \cite{feng2024more,pal2024gemini,stribling2024model}.  GPT-4 specifically has been found to exhibit limitations in processing visual information alongside text for complex reasoning tasks \cite{feng2024more}. LLMs also demonstrate variable performance in medical VQAs, with significant deficits in complex reasoning once again being reported, especially in medical imaging \cite{pal2024gemini}. Similar findings were observed in other biomedical science exams, where GPT-4 performed poorly with figure-based questions \cite{stribling2024model}.


While the inconsistent performance profile of LLMs to reason on multi-modal VQAs has not been fully explained, it is well understood in the educational domain concerning human subjects that the layout of exam questions affects students’ performance \cite{crisp2003can}. 
Similarly in the AI context, the effectiveness of an LLM's output depends on the quality and structure of the prompt. The manner in which a prompt is constructed can either effectively focus the attention of the LLM on relevant information or divert it by introducing distractions that significantly affect response accuracy. The relative position of words \cite{liu-etal-2024-lost,lu2021fantastically}, the position of an object in an image \cite{wang2024eliminating}, minor changes in wording or phrasing \cite{gao2018black,garg-ramakrishnan-2020-bae,leidinger2023language}, the order of instructions \cite{chu2024better}, and even the length of the prompt \cite{levy2024same} can all influence the accuracy of responses. These variables are magnified even more in the context of multi-modal reasoning tasks where modalities are fused and can presented to the LLMs using different strategies which are opaque to the users, and their influence on response accuracies is unknown.
Therefore, a significant challenge currently exists in determining the most effective way to construct multi-modal prompts for optimising reasoning to produce correct responses. Addressing these challenges is especially relevant for exam questions given the expansive research from the educational sector and the number of multi-modal benchmarks that are designed as exam-like questions \cite{lu2022learnexplainmultimodalreasoning,chen-etal-2024-m3cot,zhang2023m3exammultilingualmultimodalmultilevel}.

This study investigated how LLMs process and respond to variations in sequencing of images and text information when presented within multi-modal prompts via API calls. Recently, different strategies exploring ways to enhance the multi-modal reasoning of LLMs have started to emerge  \cite{Mitra_2024_CVPR,zhou2024imageofthoughtpromptingvisualreasoning,zheng2023ddcotdutydistinctchainofthoughtprompting,luan2024textcotzoomenhancedmultimodal,zhang2024multimodalchainofthoughtreasoninglanguage,susnjak2024end}. 
However, 
these strategies, while effective in certain constrained contexts, have tended to focus on a single modality without considering the interplay of modalities on the performance of reasoning tasks and have not conducted extensive experiments yielding findings and observations regarding the optimal structuring of multi-modal prompts. Our work extends and builds upon research suggesting that variations in text prompting, such as the relative position of words \cite{liu-etal-2024-lost,lu2021fantastically}, or the order of instructions \cite{chu2024better}, can significantly impact LLM performance. Understanding whether LLMs are influenced by the ordering of modalities within prompts is crucial for optimising multi-modal reasoning, thereby allowing for greater value extraction from these technologies across numerous domains.  
Consequently, this research sought to perform an extensive series of experiments to ascertain if the sequence of input modalities influences reasoning tasks, and to what extent, akin to the impact of altering instruction order in text prompts \cite{chu2024better}. This research additionally explored whether particular elements within the image and text input modalities for LLMs display sensitivity to the sequence of images and text, and if these elements can be adjusted to enhance response performance. We summarise our main contributions as follows:



\begin{enumerate}
    \item We systematically evaluated the impact of image and text prompt-sequencing on the reasoning performance of three multi-modal LLMs: GPT-4o, Gemini-1.5 Flash, and Claude-3-Haiku. Our findings demonstrate that modality sequencing significantly affects performance, particularly in complex reasoning tasks allowing us also to also speculate about the underlying modality fusion mechanisms across these models and their observed modality biases. 
    
    \item We identified specific attributes within image and text modalities that exhibit higher sensitivities to sequencing. The results indicate that different reasoning tasks benefit from distinct sequencing strategies. 
    
    \item  Based on our findings we propose practical guidelines for constructing multi-modal prompts that require complex reasoning. 
\end{enumerate}

\section{Related Work}

Reasoning can be defined as the cognitive process of drawing inferences or conclusions from premises, evidence, or observations, involving the systematic application of logical principles to analyse information, solve problems, and make decisions \cite{Johnson-Laird2010mental}. Reasoning encompasses both deductive methods, where conclusions necessarily follow from given premises, and inductive approaches, where generalisations are formed from specific instances. The assertion that reasoning is a genuine emergent behaviour in LLMs is contentious in academic literature \cite{Mitchell2024debates,Mitchell2023challenge}. Emergent abilities within LLMs have been defined as capabilities present in larger but not smaller models, with reasoning being identified as one of these properties \cite{wei2022emergent} that arise as the parameter size of language models has grown. However, recent investigations \cite{mialon2023gaia,mialon2023augmented} suggest that current LLMs find it challenging to tackle intricate reasoning tasks that humans handle with relative ease, lacking profound understanding and instead relying on superficial pattern recognition or dataset biases. Studies \cite{lecun2022path,mialon2023augmented} also argue that contemporary LLMs are confined to intuitive, reflexive tasks, rather than those necessitating logical and deliberate analysis associated with true higher-level reasoning, while others \cite{Kambhampati2024can} assert that LLMs cannot genuinely reason or plan at all, but only appear to do so. Additional research \cite{west2023generative,mcintosh2024inadequacy} further contends that the impressive generative capabilities of LLM-based systems do not reflect true understanding, but are merely a function of word prediction.

Irrespective of whether the reasoning ability exhibited by LLMs is a truly emergent property or a form of pattern-matching mimicry, this ability has been found to generalise and therefore be useful in solving many reasoning tasks \cite{Chang2024survey,Wang2024software}, thereby giving rise to the development of strategies aiming to maximise their reasoning effectiveness even further. The most recognised way of improving LLM reasoning through prompting is the Chain-of-Thought (CoT) \cite{wei2022chain} prompting technique “\textit{Let’s think step by step...}”, which has proven effective in enhancing zero-shot and few-shot capabilities \cite{wei2022chain,kojima2022large,yu2023betterchainofthoughtpromptingstrategies}. In LLMs, this method mirrors the cognitive process of breaking down problems into manageable steps, allowing the model to process each step sequentially in a linear fashion, ultimately leading to a conclusive answer\cite{wang-etal-2023-towards}.
Recent advancements in multi-modal reasoning for LLMs have focused on enhancing CoT methods to address challenges like their weak spatial reasoning, localisation awareness \cite{Ranasinghe_2024_CVPR,chen2024spatialvlmendowingvisionlanguagemodels}, and high-resolution image interpretation \cite{luan2024textcotzoomenhancedmultimodal}. 
The upcoming challenge in reasoning complexity lies in further enhancing the abilities of LLMs to reason across various input modalities, including text and image elements, and eventually other multimedia types as well\cite{susnjak2024end}. Research in this area is nascent but has repeatedly shown the need to devise improved means for LLMs to perform multi-modal reasoning more reliably \cite{pal2024gemini,stribling2024model}. While multi-step reasoning follows a sequential approach to draw conclusions as exemplified by CoT approaches, multi-hop reasoning, however, requires making several inferential jumps among unconnected data points or different modalities to form a coherent answer which presents a significant degree of difficulty for transformer-based architectures which are unable to iteratively plan and refine their responses.


\subsection{Multi-modal Prompting Techniques}
Several studies have focused on improving and addressing LLM's challenges within the vision modality. Techniques such as Compositional Chain-of-Thought (CCoT) \cite{Mitra_2024_CVPR} use scene graph-based prompting to achieve this while Image-of-Thought (IoT) \cite{zhou2024imageofthoughtpromptingvisualreasoning} extracts visual rationales in a step-by-step manner. Meanwhile, TextCoT \cite{luan2024textcotzoomenhancedmultimodal} divides images into global and local regions to assist with reasoning, while Duty-Distinct Chain-of-Thought (DDCoT)  \cite{zheng2023ddcotdutydistinctchainofthoughtprompting}  employs a two-stage framework to separate reasoning roles for visual and language modalities.  Multi-modal Chain-of-Thought (MCoT)\cite{zhang2024multimodalchainofthoughtreasoninglanguage} improves multi-modal reasoning by initially partitioning LLM responsibilities into reasoning and recognition before integrating vision information within smaller models. Although these models examine the interaction between modalities, they treat them as distinct components that can be processed independently. These strategies, while effective in certain contexts, have not considered how the sequencing of modalities affects reasoning performance.

\subsection{Image Sequencing}
In human behaviour, the \textit{primacy} effect suggests that individuals are more likely to recall information presented at the beginning of a sequence \cite{asch1946forming}, in contrast to the \textit{recency} effect, which implies a contrary bias towards information at the end of a sequence \cite{baddeley1993recency}. Both the primacy and recency effects have been demonstrated to exist within LLMs \cite{liu-etal-2024-lost,lu2021fantastically,wang-etal-2023-primacy,zhang2023exploring,eicher2024compensatory}; however, these have not been comprehensively studied and explored in the context of multi-modal LLMs and reasoning tasks. Vendors \cite{googleimageunderstanding,anthropic2024claude,OpenAI2023community} of large commercial LLMs have tended to advise that in cases involving prompts with images, there is a primacy effect that impacts performance\footnote{Both Google \cite{googleimageunderstanding} and Anthropic \cite{anthropic2024claude} recommend placing the image first to achieve the best results. The OpenAI community pages \cite{OpenAI2023community} have a more nuanced recommendation, suggesting that placing the image first often helps the LLM in understanding the tasks and framing the problem.}. For general tasks where the image is the focus, this logic makes sense; however, for reasoning tasks where key instructions are often in a dedicated question component, this may not hold true. To the best of our knowledge, there is little information on why this is recommended or evaluations on different types of tasks for image position.


\subsection{Multi-modal Fusion Strategies and Positional Bias}

The architectural foundation of LLMs plays an important role in how the sequencing of information is processed. The architecture of LLMs is based on transformers \cite{vaswani2017attention}, which use attention mechanisms to assign varying weights to input data based on context. This involves multiple self-attention layers running in parallel, enabling LLMs to simultaneously focus on different aspects or relationships between tokens in the input sequence. These patterns are learned through training or further refined via fine-tuning. For text, transformers use tokenised word representations with positional encoding to maintain sequence order, which is critical for understanding context and syntax  \cite{vaswani2017attention,devlin2018bert}. 
The integration of multiple modalities such as text and images within LLMs requires effective fusion strategies to enable coherent understanding and reasoning. Fusion strategies determine how information from different modalities is combined and processed within a model during pre-training. The primary fusion strategies are \textit{early fusion}, \textit{late fusion}, and \textit{hybrid fusion}, each with distinct implications for multi-modal prompting and complex reasoning tasks \cite{zhao2024deep}.

Early fusion is also known as input-level fusion. This approach involves integrating different modalities at the initial stage by converting them into a unified token representation before feeding them into the model \cite{tsai2019multimodal,team2024chameleon}. In transformer-based architectures, this typically means embedding images and text into a shared embedding space and concatenating them into a single input sequence from the beginning \cite{lu2019vilbert,li2021align,team2024chameleon}.
This strategy allows the model to learn cross-modal interactions from the outset. Cross-modal interactions are the relationships and dependencies between different data modalities (like text, images, and audio) in multi-modal machine learning, where  information from one modality influences or complements another to enhance a model's overall understanding and reasoning. This capability ultimately enables the model to capture fine-grained relationships between modalities leading to more seamless reasoning and generation across modalities \cite{team2024chameleon}. While early fusion may be advantageous for tasks requiring deep integration of modalities such as visual question answering and image captioning; it can introduce challenges in processing efficiency and scalability, especially when dealing with high-dimensional data like images since the model must handle large input sequences, which can increase computational complexity and memory requirements \cite{gan2020large,team2024chameleon}.

In contrast, late fusion, otherwise referred to as decision-level fusion, processes each modality independently through separate sub-networks and combines their outputs after feature extraction \cite{baltruvsaitis2018multimodal}. This approach allows each modality to be encoded optimally for its unique characteristics without interference from others. Late fusion is effective when modalities contribute independently to the final decision or when cross-modal interactions are less critical. It offers computational advantages by enabling parallel processing and reducing the complexity associated with handling combined input sequences. However, late fusion may not capture nuanced cross-modal relationships essential for tasks that require integrated reasoning across modalities. The separation of modalities can limit the model's ability to perform complex reasoning dependent on the interplay between different types of information \cite{liang2021multibench}.

Hybrid fusion strategies combine elements of both early and late fusion to leverage their respective strengths \cite{baltruvsaitis2018multimodal}. In hybrid fusion, certain modalities are fused early to capture essential interactions, while others are integrated at later stages \cite{dou2022empirical}. This approach provides flexibility in modelling cross-modal relationships at different levels of abstraction \cite{singh2022flava}. Hybrid fusion is particularly beneficial for complex tasks requiring layered reasoning across modalities. Layered reasoning across modalities is the hierarchical integration and interpretation of information from different data types at multiple levels of abstraction within a model, enabling it to capture complex interactions by progressively combining multi-modal data through successive layers. This ability balances the need for deep integration of specific modalities with the efficiency of processing others independently \cite{akbari2021vatt}.

Across different input modalities, recent research \cite{wang2024eliminating,leidinger2023language,garg-ramakrishnan-2020-bae} has shown that both text and images are susceptible to positional bias, attributing this to the manner in which causal attention and relative positional encoding operate in most LLMs \cite{wang2024eliminating}, which is likely an artefact of pre-training \cite{peysakhovich2023attention}. Positional bias can also extend to the sequence of instructions which significantly impacts LLM performance \cite{chu2024better}, suggesting that the placement of modalities and the order of instructions are crucial for effective multi-modal reasoning. This aligns with the work of \cite{gao2018black, garg-ramakrishnan-2020-bae,leidinger2023language} which identified that even minor changes in wording or phrasing can affect performance. 

Large vendors of proprietary LLMs typically do not disclose the implementation details of their commercial multi-modal models which can make it challenging to know how to optimise prompts for the most accurate reasoning responses. Even though different modalities may be processed separately initially, the order in which they are presented in the prompt can still influence a model's reasoning performance due to the mechanics of positional encoding and attention in transformer architectures. In early fusion architectures—where modalities are integrated at the input level into a unified token sequence—modality sequencing has a significant impact because position directly affects how the model attends to and integrates information. Hybrid fusion and late fusion systems, which process modalities independently before combining them at later stages, may exhibit less sensitivity to modality order; however, prompt design and sequencing can still affect performance by influencing how information is integrated during fusion. Therefore, understanding these internal mechanics across different fusion strategies allows for more effective prompt design and optimisation of multi-modal LLMs for complex reasoning tasks.


\subsection{Research questions}

Recent literature is collectively beginning to converge towards investigations that seek to uncover strategies to optimise prompts for maximising LLM performance and reasoning. While existing research has mostly tended to focus on enhancing performance gains within a single modality (text), in cases where multi-modal information was considered, the studies have typically overlooked the impact of information sequencing in multi-modal contexts and how different and unknown multi-modal fusion strategies may be a confounding factor that affects responses.  Therefore, our research has aimed to bridge this gap by examining how the sequencing of images and text affects LLM performance in reasoning tasks. 
To that end, this study's guiding research questions are:

\begin{itemize}
    \item RQ1: To what extent does the sequencing of image and text modalities in prompts affect the reasoning performance of multi-modal LLMs having different multi-modal fusion strategies, across different benchmark datasets and question types
    
    \item RQ2: How do specific attributes of questions, such as nested structure, subject domain, and complexity, interact with modality sequencing to influence LLM performance, and how does this vary across different LLM models?
    
    \item RQ3: To what degree is the impact of modality sequencing on LLM performance attributable to the order of information presentation rather than the inherent properties of different modalities, and how can these insights be applied to optimise multi-modal prompt construction?
\end{itemize}

\section{Methodology}

We designed a series of experiments on two benchmark datasets detailed in the subsequent section to address our research questions. 

\subsection{Datasets}
Our evaluations used two recently developed multi-modal multiple-choice reasoning benchmarks for LLMs, namely M3Exam\cite{NEURIPS2023_117c5c86} and M3COTS\cite{chen-etal-2024-m3cot}. These benchmarks were developed with questions that integrate visual and textual information and were thus selected in our experiments due to their ability to present models with both complex and demanding reasoning tasks from multiple modalities.

\subsubsection{M3Exam Dataset}

M3Exam\cite{NEURIPS2023_117c5c86} offers a diverse range of real exam questions across various educational levels. For our evaluation, we selected the multi-modal English question set which contains 795 questions across 4 overarching subjects (social-science, natural-science, language, math), 11 subcategories, and 3 educational levels (elementary, middle, and high school) in the USA. The average word count across the questions and background information is approximately 95 words. 

\begin{figure}[htbp]
    \centering
        \centering
        \includegraphics[width=0.5\linewidth]{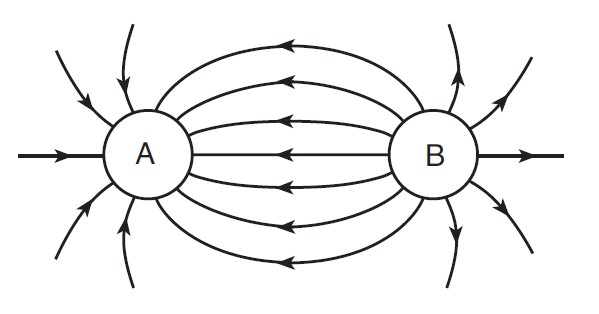}
        \caption{ M3Exam example question 5}
        \label{fig:M3ExamExampleQuestion5}
\end{figure}

The M3Exam dataset structures each question in JSON format, dividing it into three key parts:  \texttt{background\_description} which provides additional context in some cases, the \texttt{question\_text} which contains the actual questions, and \texttt{options} which represents the multiple-choice responses. Image elements can be dispersed across all three elements and, sometimes in multiple places per question which further amplifies the complexity of questions. An example of an exam question with three components can be seen in Figure \ref{fig:M3ExamExampleQuestion5}, with guidance suggesting that the image component be placed in the \texttt{question\_text} section of the overall question. This particular question does possess an empty  \texttt{background\_description} component.

\begin{small}
\begin{verbatim}
    
{
    "background_description": [],
    "question_text": "The diagram below represents the electric field surrounding two charged
    spheres, A and B.\n\n(image)[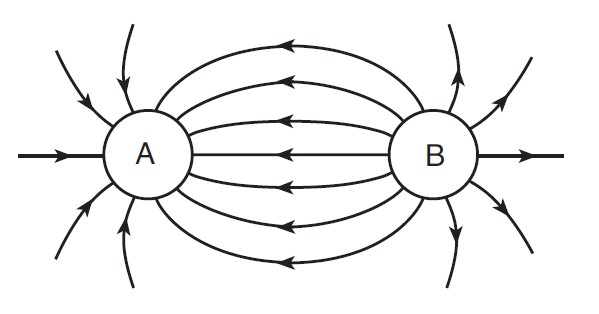]\n\nWhat is the sign of the charge of each 
    sphere?",
    "options": [
        "(1) Sphere A is positive and sphere B is negative.",
        "(2) Sphere A is negative and sphere B is positive.",
        "(3) Both spheres are positive.",
        "(4) Both spheres are negative."
    ]
}
\end{verbatim}
\end{small}

\begin{figure}[ht]
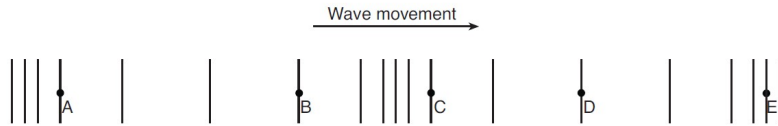
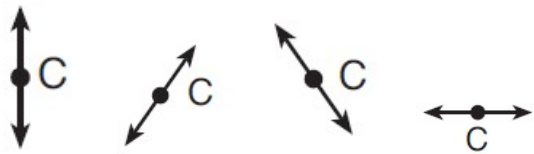

    \centering
    \begin{subfigure}[b]{0.65\textwidth}
        \centering
        \includegraphics[width=\textwidth]{image-10.jpg}
        \caption{Image-10}
    \end{subfigure}

    \vspace{0.15cm} 
    \begin{subfigure}[b]{0.1\textwidth}
        \centering
        \includegraphics[width=\textwidth]{image-11.jpg}
        \caption{Image-11}
    \end{subfigure}
    \begin{subfigure}[b]{0.12\textwidth}
        \centering
        \includegraphics[width=\textwidth]{image-12.jpg}
        \caption{Image-12}
    \end{subfigure}
    \begin{subfigure}[b]{0.12\textwidth}
        \centering
        \includegraphics[width=\textwidth]{image-13.jpg}
        \caption{Image-13}
    \end{subfigure}
    \begin{subfigure}[b]{0.12\textwidth}
        \centering
        \includegraphics[width=\textwidth]{image-14.jpg}
        \caption{ Image-14}
    \end{subfigure}
    
    \caption{Set of images from the M3Exam dataset showing a complex set of image arrangements.}
    \label{fig:m3exam_examples}
\end{figure}

Since visual elements can be distributed across the three elements at the same time, the complexity arising from multiple multi-modal inputs can be significant for some exam questions. An example is given in Figure \ref{fig:m3exam_examples} where an image component is allocated to the \texttt{background\_description} component, while a further four images are allocated to each of the four answer options. The JSON structure of the question is depicted below showing image placeholders, e.g. denoted as \texttt{(image)[image-x.jpg]}. Overall, the questions in the dataset range from having 1 to a maximum of 5 images, averaging 1.2 images per question in the dataset. 


\begin{small}
\begin{verbatim}
{
    "background_description": [
        "A longitudinal wave moves to the right through a uniform medium, as shown 
        below. Points A, B, C, D, and E represent the positions of particles of the 
        medium.\n\n(image)[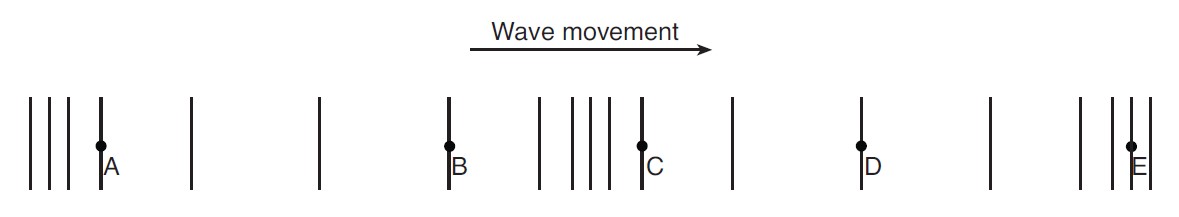]"
    ],
    "question_text": "Which diagram best represents the motion of the particle at 
    position C as the wave moves to the right?",
    "options": [
        "(1) (image)[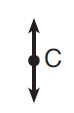]",
        "(2) (image)[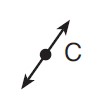]",
        "(3) (image)[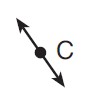]",
        "(4) (image)[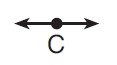]"
    ]
}
\end{verbatim} \label{m3exam_example}
\end{small}

\begin{wrapfigure}{r}{0.5\textwidth}
\includegraphics[width=0.8\linewidth]{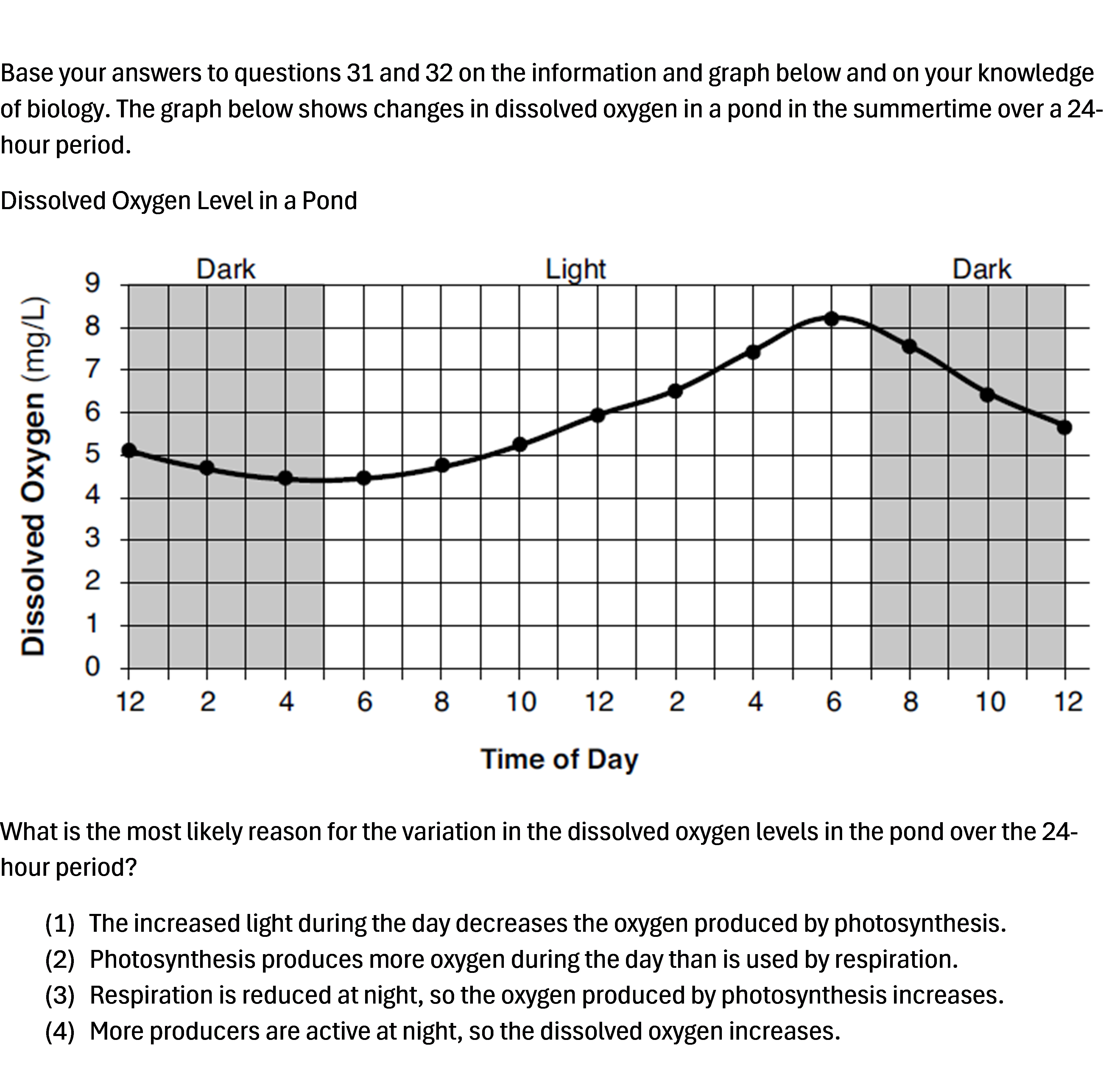} 
\caption{ An example of a reconstructed M3Exam question.}
\label{fig:M3ExamExampleQuestion}
\end{wrapfigure}

An example of a complete reconstructed exam question is shown in Figure \ref{fig:M3ExamExampleQuestion}, where the image elements are situated in the background/context portion of the question. In terms of image placement, 87\% of images in the dataset are situated inline within \texttt{background\_description} or  \texttt{question} component and 6\% within the options component. Meanwhile, 7\% of the images appear at the start of the question within the \texttt{question\_text} component. Given that the exam questions are deconstructed in the raw data, they lend themselves well to modifying the order in which the modalities are presented to the LLMs through the API calls. Further, since these are actual exam questions used in the US education sector, their layout is assumed to be optimised for student understanding. M3Exam has been used to evaluate models focused on different languages \cite{nguyen2023seallms} and culture-related tasks \cite{Liu2024IsTA}, making it a versatile benchmark.  

\subsubsection{M3COTS Dataset}

\begin{figure}[htbp]
    \centering
    \begin{subfigure}[b]{0.48\linewidth}
        \centering
        \includegraphics[width=\linewidth]{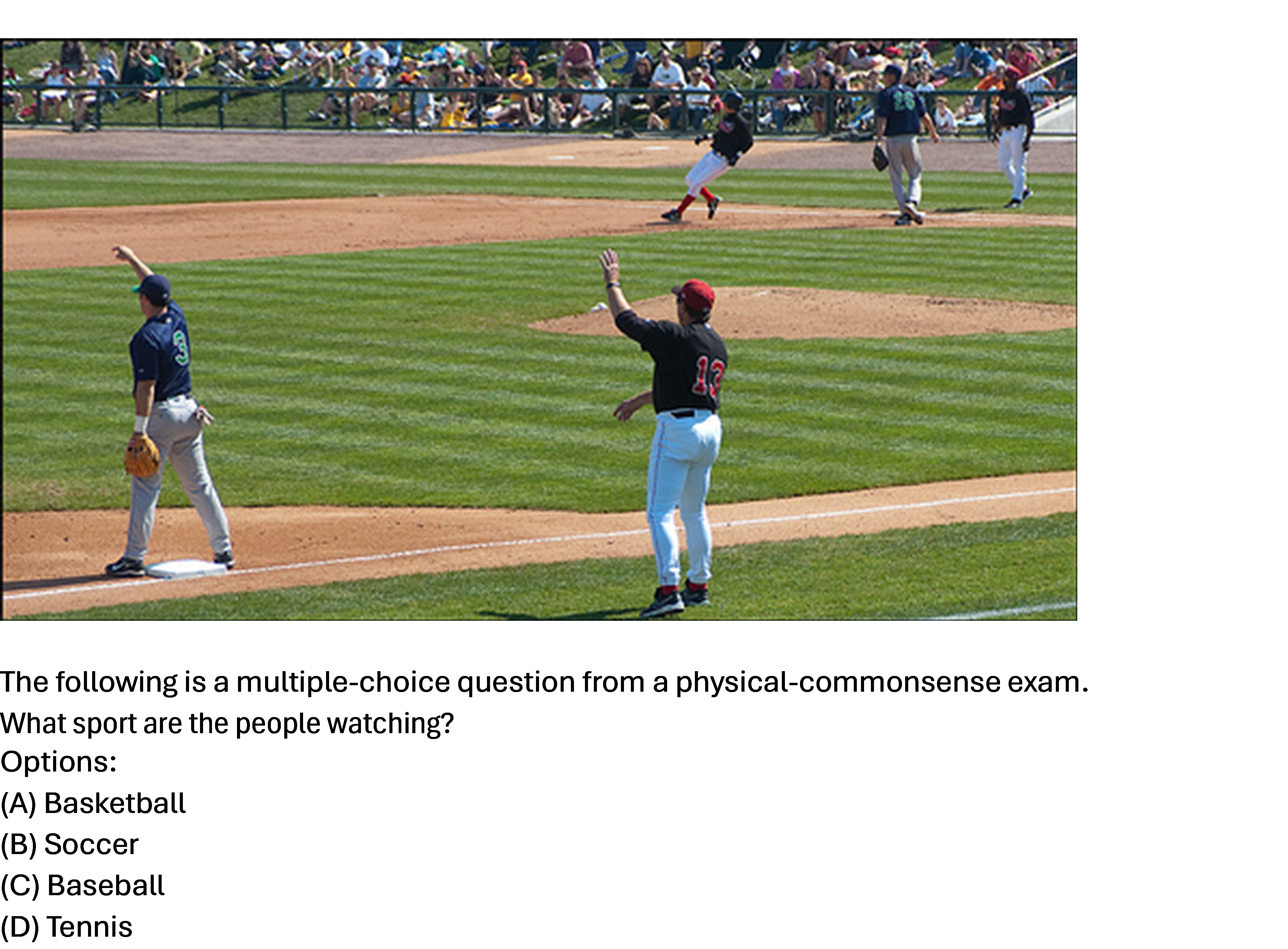}
        \caption{Example - images containing only visual content}
        \label{fig:M3COTSExampleQuestion2}
    \end{subfigure}
    \hfill
    \begin{subfigure}[b]{0.48\linewidth}
        \centering
        \includegraphics[width=\linewidth]{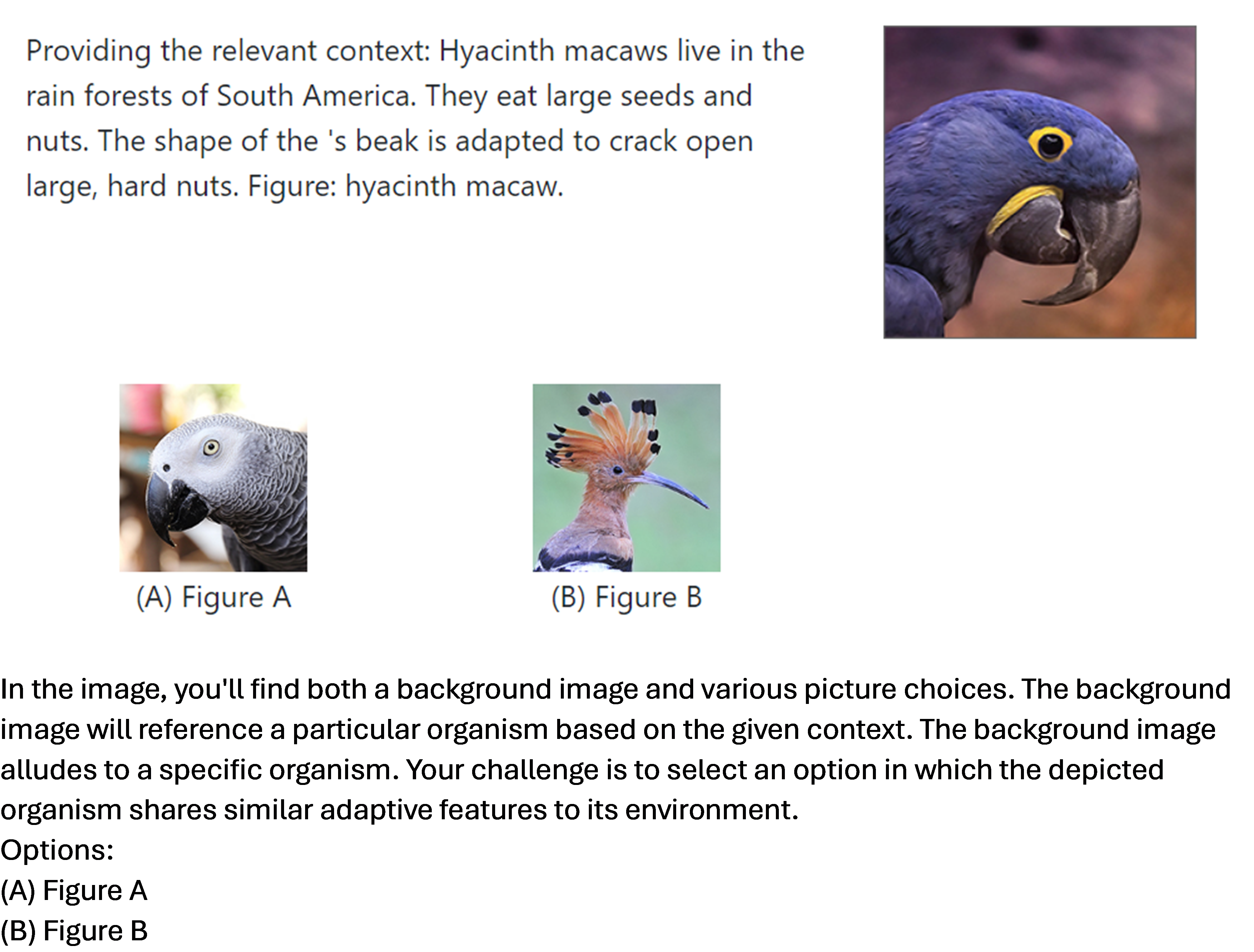}
        \caption{Example - images containing text and visual content}
        \label{fig:M3COTSExampleQuestion1}
    \end{subfigure}
    \caption{Typical text/image layouts across M3COTS dataset questions with the image first.}
\end{figure}

The second dataset used in this analysis is M3COTS \cite{chen-etal-2024-m3cot}. M3COTS features a selection of questions specifically chosen to challenge visual reasoning and multi-step reasoning across multiple subjects. The dataset includes science topics from the ScienceQA dataset \cite{lu2022learnexplainmultimodalreasoning}, mathematics questions from MATH \cite{hendrycksmath2021} and the Sherlock \cite{hessel2022abduction} datasets, intended to test common-sense abductive reasoning beyond the literal image content. For our evaluation, we selected a random sample of 2,318 questions (20\% of the dataset) spanning 3 domains, 9 subjects, and 92 question types.
In this dataset, each question includes only one image as opposed to M3Exam which is a significant reduction in complexity. 
The average word count across the questions, background information, and options is approximately 45 words. Ten percent of the images contain only visual content, 65 percent consist of a combination of images and text, and 25 percent feature text exclusively. Example questions from M3COTS are shown in Figures \ref{fig:M3COTSExampleQuestion2} and \ref{fig:M3COTSExampleQuestion1}. Note that as seen in Figure \ref{fig:M3COTSExampleQuestion1}, while each question in this dataset may be accompanied by only one image in the raw format, an image may however embed multiple images distinct and as well as text within a single visual.
Similarly to the M3Exam dataset, M3COTS structures each question in JSON format, dividing it into three key parts: \texttt{context} which provides additional background in some cases, the \texttt{question} component which contains the actual question, and \texttt{choices} which represents the multiple-choice responses. The images are not directly referenced in the \texttt{context},  \texttt{question} or \texttt{choices}. The example JSON structure of a M3COTS question can be seen below.
\begin{small}

\begin{verbatim}
{
    "image": physics-26.png,
    "context": "Select the better answer.",
    "question": "Which property do these two objects have in common?",
    "choices": [" 
                 "(A) sticky", 
                 "(B) yellow"
                ]
}
\end{verbatim}
\end{small}

The diverse range of domains and source datasets makes M3COTS a suitable benchmark for evaluating LLMs. The source  and M3COTS dataset has been investigated extensively in research. CoT \cite{wei2022chain} prompting has proven to be the most effective technique outperforming Direct Prompting, Description-based CoT (Desp-CoT)\cite{wu2023role}, and Compositional CoT (CCoT) \cite{Mitra_2024_CVPR}.

\subsection{LLM Models}

We selected three popular commercial models for our experiments: ChatGPT-4o, Claude-3.5, and Gemini-1.5 Flash.
ChatGPT-4o was developed by OpenAI and introduced in 2023, is characterised by a large parameter count and extensive context length. These features enable sophisticated multi-modal interactions and complex reasoning tasks. Claude-3.5 Haiku was produced by Anthropic, and is recognised for its speed and compact design. This model provides an ideal contrast to larger, more computationally intensive models like ChatGPT-4o, offering insights into the trade-offs between model size and response latency. Lastly, Gemini-1.5 Flash is Google's model and is regarded as another ``lightweight'' model optimised for speed and efficiency, complementing the other selections by focusing on streamlined performance.

The three models, with their varying capabilities and architectural designs, collectively provide a comprehensive overview of the current landscape in large-scale AI computations. The decision to focus on larger LLMs stems from existing studies\cite{wei2022chain}, which suggest that the capability for Chain-of-Thought (CoT) reasoning may emerge in language models at a certain scale, specifically over 100 billion parameters. All models were accessed via their respective APIs, hosted on platforms capable of supporting extensive AI operations, thereby ensuring reliable and consistent performance throughout our studies.

The experiments were conducted in a zero-shot fashion, ensuring that the models were not exposed to any examples prior to testing. We employed variations of CoT's prompts \cite{wei2022chain}, and all testing was conducted using greedy decoding at a temperature setting of 0.1. Our experiments used standard models without any fine-tuning to focus on the models' behaviour under direct interaction, which is the most common approach users take when engaging with language models. We opted not to sample multiple responses or perform self-consistency-based re-ranking \cite{wang2023selfconsistencyimproveschainthought}, as these methods significantly increase operational costs and may not be practical in many scenarios related to our datasets.

In this research, the focus was on examining the relative performance of the chosen LLMs across different image and text input configurations. Therefore, the primary aim was not to achieve maximal state-of-the-art performance, but rather to understand how these models behave with changes to the sequencing configuration of the text and image inputs. 

\subsection{Experimental design}

This study conducted a series of experiments to evaluate how the sequencing of image and text modalities in prompts affects the multi-hop reasoning performance of multi-modal LLMs, structured around four primary setups: (1) \textit{Image-Text Sequence Variation}, which examined the effects of different sequencing orders (Image First, Text First, and Interleaved) on model performance across two datasets; (2) \textit{Attribute-Based Sequencing Analysis}, which investigated how specific dataset attributes—such as image type, prompt length, and question complexity—influence the model’s sensitivity to sequencing; (3) \textit{Image Versus Instructions Analysis}, aimed at determining whether the impact of sequencing is due to the image placement or the sequence of instructions by converting visual elements into text; and (4) \textit{Prompt Priming for Relationship Analysis}, which explored whether priming the model to prioritise a specific modality alters its reasoning process, irrespective of the initial sequencing. Table \ref{tab:experimental_design_summary} summarises the entire  experimental design, which is explained in further detail below.

\begin{table}[h]
\centering
\caption{Overview of the experimental design}
\fontsize{8pt}{10pt}\selectfont
\begin{tabular}{p{2cm}p{3cm}p{2cm}p{3cm}p{4cm}}
\hline
\textbf{Experiment} & \textbf{Description} & \textbf{Configurations} & \textbf{Variables Analysed} & \textbf{Hypothesis} \\ \hline

\textbf{Image-Text Sequence Variation} & 
Evaluates effect of sequencing on model performance & 
\begin{tabular}[c]{@{}l@{}} 
Image First (IF) \\ 
Text First (TF) \\ 
Interleaved (IN)
\end{tabular} & 
\begin{tabular}[c]{@{}l@{}} 
Impact of sequencing \\ 
on reasoning performance
\end{tabular} & 
\begin{tabular}[c]{@{}l@{}} 
Sequencing affects LLM performance \\ 
with the best configuration depending \\ 
on the dataset
\end{tabular} \\ \hline

\textbf{Image-Text Sequence: Attribute-Based Analysis} & 
Investigates whether the relationships or trends observed in the overall dataset hold for each of the attributes & 
\begin{tabular}[c]{@{}l@{}} 
Image First (IF) \\ 
Text First (TF)
\end{tabular} & 
\begin{tabular}[c]{@{}l@{}} 
\textbf{Attributes:} \\ 
- Image Type \\ 
- Prompt Length \\ 
- Difficulty Levels \\ 
- Question Types
\end{tabular} & 
\begin{tabular}[c]{@{}l@{}} 
Attribute should follow the same\\ 
pattern as observed in the overall dataset
\end{tabular} \\ \hline

\textbf{Image vs Instructions Analysis} & 
Determines if sequencing impact is due to image placement or instruction sequence & 
\begin{tabular}[c]{@{}l@{}} 
Image First (IF) \\ 
Text First (TF)
\end{tabular} & 
\begin{tabular}[c]{@{}l@{}} 
Impact of sequencing on \\ 
extracted text from images \\ 

\end{tabular} & 
\begin{tabular}[c]{@{}l@{}} 
The sequence of instructions \\ 
affects performance, independent \\ 
of image placement
\end{tabular} \\ \hline

\textbf{Prompt Priming for Relationship Analysis} & 
Explores effect of priming on reasoning process & 
\begin{tabular}[c]{@{}l@{}} 
Image First (IF) \\ 
Text First (TF)
\end{tabular} & 
\begin{tabular}[c]{@{}l@{}} 
Priming to prioritise \\ 
image or text processing
\end{tabular} & 
\begin{tabular}[c]{@{}l@{}} 
Priming the LLM to focus on a specific \\ 
modality influences its ability \\ 
to answer questions accurately
\end{tabular} \\ \hline

\end{tabular}
\label{tab:experimental_design_summary}
\end{table}

\subsubsection{Image-Text Sequence Variation}

This experiment investigated the zero-shot multi-modal reasoning, where the model was tasked with predicting an answer \(a\) to a prompt that included a textual query \(q\) and an image \(x\), without having been exposed to similar tasks during training. The model was required to analyse both the visual content in \(x\) and the information in \(q\), integrating these inputs to generate a correct response. The experiment was specifically designed to evaluate how the sequence and integration of textual and visual inputs, as structured within the API calls,  affect the model's reasoning capabilities. 

\begin{figure}[htbp]
    \centering
        \centering
        \includegraphics[width=0.75\linewidth]{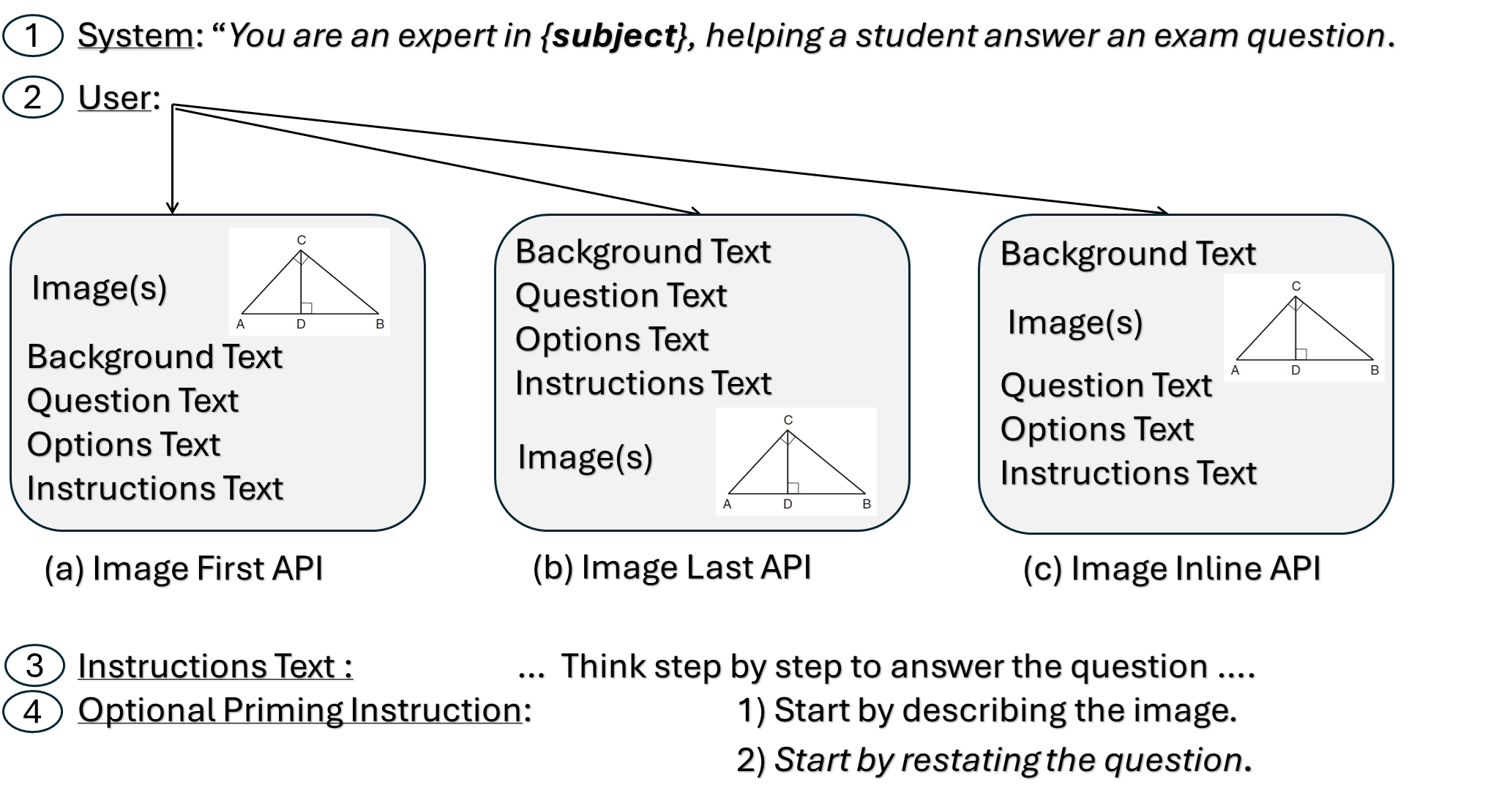}
        \caption{Example of the structure of the API calls containing the prompts for different experimental configurations.}
        \label{fig:apis}
\end{figure}

Each of the three models' API's encodes information in a similar manner where a set of parameters along with a prompt is sent to the model as depicted in Figure \ref{fig:apis}. The prompt was composed of information from different \textit{roles}, which defined the context and purpose of each part of the message. For this experiment, the prompt consisted of messages from two key roles: \textit{system} and \textit{user}. The \textit{system} message sets the overall tone and controls how the model should respond. In this experiment, we used a fixed template for the system message:  \textit{"You are an expert in \{subject\}, helping a student answer an exam question."}.  This message remained constant across all configurations, ensuring a consistent context for the model’s responses. The second role in our prompt was the \textit{user} role, which represents the input or question provided to the model. The user role contained blocks of content that can include either images or text. Since our experiments tested how the order of these content blocks (text and images) affects the model’s performance, we varied the sequence in which the content blocks were presented to the LLM.  We tested three configurations: \textit{Image First}, \textit{Text First}, and \textit{Interleaved}, to determine their impact on the model's performance. The response \(a\) generated by the model under each configuration is defined as follows:

\begin{itemize}
    \item \textbf{Image First (IF):} The model processes the image \(x\) before the text \(q\), represented by the function \(f_{\text{IF}}\). 
    \[a_{\text{IF}} = f_{\text{IF}}(x, q)\]
    \item \textbf{Text First (TF):} The model processes the text \(q\) before the image \(x\), represented by the function \(f_{\text{TF}}\).
    \[a_{\text{TF}} = f_{\text{TF}}(q, x)\]
    \item \textbf{Interleaved (IN):} The model processes blocks of text (\(q_1, q_2, \ldots, q_n\)) interspersed with the image \(x\), integrating these inputs in sequence, represented by the function \(f_{\text{IN}}\).
    \[a_{\text{IN}} = f_{\text{IN}}(q_1, x, q_2, \ldots, q_n)\]
\end{itemize}

The above experiments were translated into API calls in the formats depicted in Figure \ref{fig:apis} and comprised four components and steps. In step 1, the LLM is invoked to assume a subject expert persona for each respective field associated with a given question. This was then followed by step 2 which varied the sequencing of the image and textual components of the questions. In step 3, the LLM is given a standard CoTs instruction to \textit{ "Think step by step to answer the question, ..."}  across all configurations. In experiments involving prompt priming, a further instruction was appended to this prompt as seen in step 4 which could take either the instruction to focus attention on the image or the question.

\textbf{Hypothesis:} "The sequence in which images and text are presented significantly affects the ability of a LLM to accurately answer multiple-choice questions. We hypothesise that:

\begin{itemize}
    \item For the M3Exam dataset, where images are interleaved with text, the \(f_{\text{IN}}\) configuration will yield the best performance.
    \item For the M3COTS dataset, where images are typically presented before the question, the \(f_{\text{IF}}\) configuration is expected to yield the best performance.

    \end{itemize}

\subsubsection{Image-Text Sequence: Attribute-Based Analysis}

In these experiments, we analysed how varying attributes within the dataset—such as the type of image (image, text, or a mixture of both), prompt length, difficulty levels, and question types—affect the model's performance and sensitivity to sequencing. The goal was to examine whether the trends observed in the overall dataset hold for each of the attributes.
\begin{itemize}
\item \textbf{Image Type:} The model's performance is evaluated based on different types of images—purely visual, text-based, and mixed images.
\item \textbf{Prompt Length:} Various lengths of prompts are tested to observe how the length of the text portion affects the model's accuracy and reasoning capabilities.
\item \textbf{Difficulty Levels, and Question Types:} The experiment evaluates how different difficulty levels, and question types within the dataset influence the model's performance.
\end{itemize}

To quantify the impact of these attributes on the LLM's reasoning process, we introduce the following formula:   \[
    a_{\text{AB}} = f_{\text{AB}}(c, d, \text{Attr})
    \]
where \(\text{Attr}\) represents the attributes of the dataset being evaluated with \(c\) and \(d\) represent the different sequencing configurations of the modalities
 
\subsection{Image-Text Sequence: Image Versus Instructions Analysis}

\begin{figure}[htbp]
    \centering
    \begin{subfigure}[b]{0.48\linewidth}
        \centering
    \includegraphics[width=\linewidth]{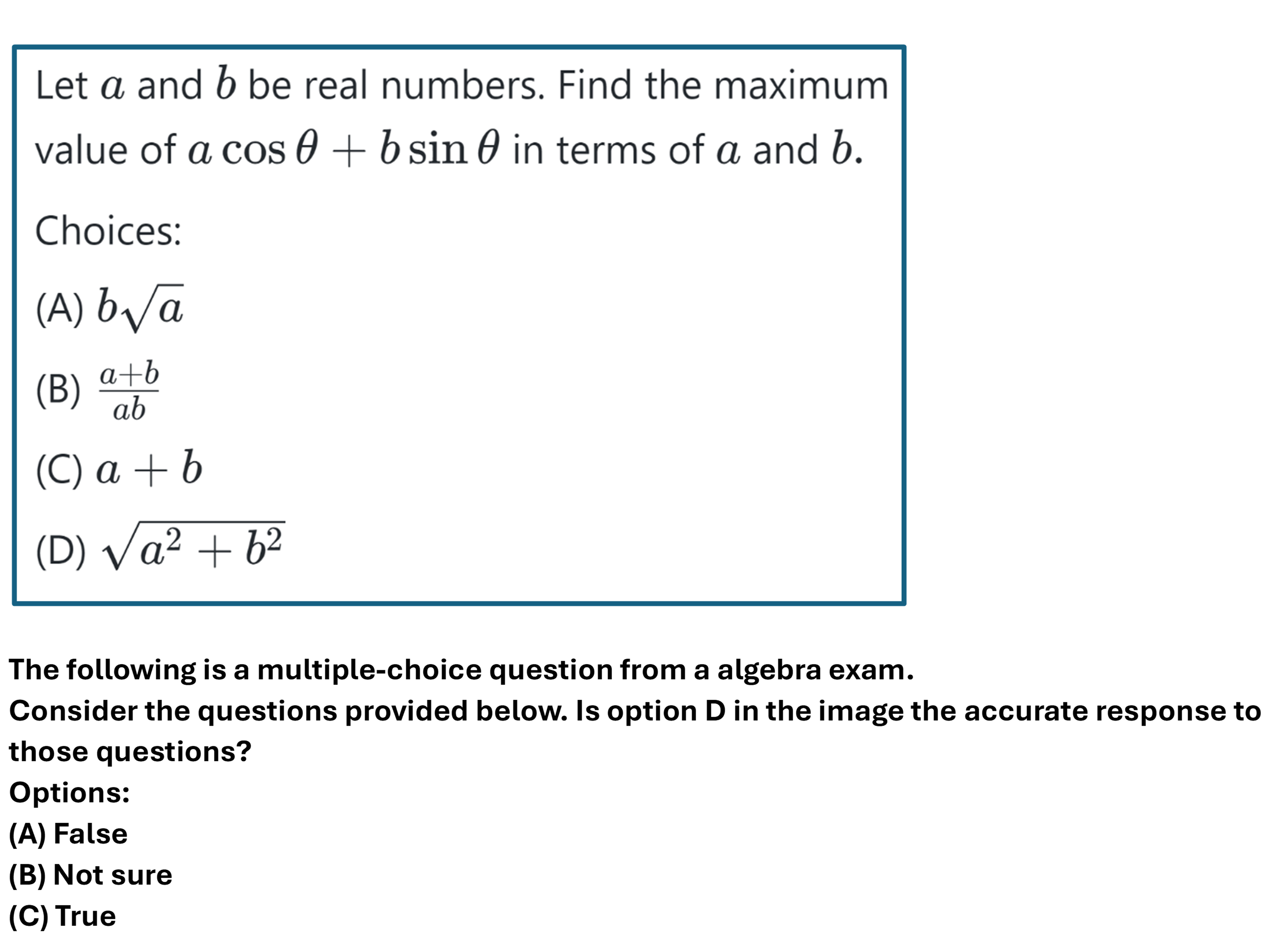}
    \caption{Example question with an image containing embedded text}
    \label{fig:mathematics-1358_withImage}
    \end{subfigure}
    \hfill
    \begin{subfigure}[b]{0.48\linewidth}
        \centering
    \includegraphics[width=\linewidth]{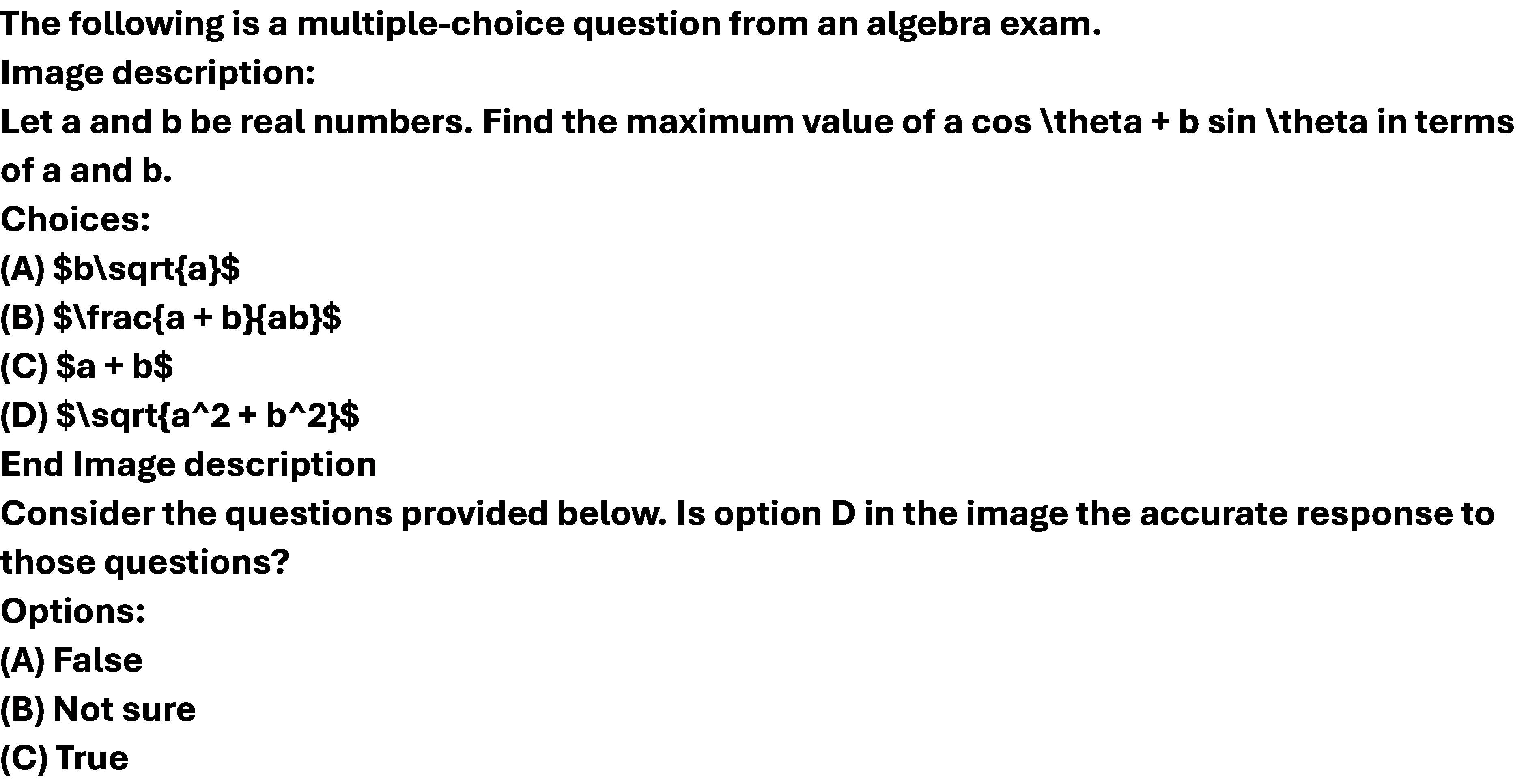}
    \caption{Identical question with textual content extracted from the image and converted to pure text}
    \label{fig:mathematics-1358_AllText}
    \end{subfigure}
\caption{Example of an image-based question converted to a pure text-based question}
    \label{fig:mathematics-1358}
\end{figure}

To determine whether the impact of sequencing is due to the placement of the image or the sequence of text-based prompting instructions, we conducted experiments on a selected sample of question types from the M3COTS dataset. These questions contained only text or embedded formulas within the images. We extracted and converted the visual content into text (referred to as \(x_{\text{TextExtracted}}\)) and ran the sequencing experiments using the text modality only. This approach allowed us to control for, and identify whether performance differences arise from the image's placement or the phrasing and sequencing of the instructions\ref{fig:mathematics-1358}.

The specific configurations being tested are:

\begin{itemize}
    \item \textbf{Image First (IF):} The model processes the extracted text from the image \(x_{\text{TextExtracted}}\) before the textual query \(q\). This is represented by the function \(f_{\text{IF}}\). 
    \[a_{\text{IF}} = f_{\text{IF}}(x_{\text{TextExtracted}}, q)\]
    \item \textbf{Text First (TF):} The model processes the textual query \(q\) before the extracted text from the image \(x_{\text{TextExtracted}}\), represented by the function \(f_{\text{TF}}\).
    \[a_{\text{TF}} = f_{\text{TF}}(q, x_{\text{TextExtracted}})\]
    
\end{itemize}

\subsection{Prompt Priming for Relationship Analysis}
We also introduced a \textit{priming} mechanism, denoted as \(p\), which was used to explicitly instruct the model to focus its attention either on the image \(x\) first or on the text query \(q\). The objective was to influence the order in which the model processed each modality in the multi-hop reasoning order, regardless of their initial presentation sequence.

\begin{enumerate}
    \item \textbf{Single Prompt—Image First Attention (IFA):} In this configuration, even though the image is presented second, the primed prompt instructs the LLM to prioritise processing and its attention on the image. The model's response is defined as \(a_{\text{IF}} = f_{\text{IF}}(x, q, p)\), where \(p\) includes specific instructions to first focus on \(x\). This priming seeks to alter the model's attention mechanisms. The prompt used is: ``Think step by step to answer the question. Start by describing the image.''

    \item \textbf{Single Prompt—Text First Attention (TFA):} In this configuration, even though the text is presented second, the primed prompt instructs the LLM to prioritise processing and its attention on the text. The model's response is defined as \(a_{\text{TF}} = f_{\text{TF}}(q, x, p)\), where \(p\) modifies the sequence to process \(q\) before \(x\), potentially reshaping the model's initial focus. The prompt used is: ``Think step by step to answer the question. Start by restating the question.''
\end{enumerate}

\textbf{Hypothesis Tested:}
\begin{itemize}
    \item The hypothesis tested is that priming the LLM to focus on a specific modality at the start of the prompt will affect the model's ability to accurately answer questions, similar to the impact observed with modality sequencing.
\end{itemize}

\subsection{Inference into the LLM Modality Fusion Strategy}

The study hypothesises that the impact of modality sequencing on model performance will vary depending on the unknown fusion strategy employed by the underlying LLMs. For early fusion models, where all modalities are processed together as a unified token sequence, we expect significant sensitivity to the order of images in the prompt. Configurations such as image-first or image-last are likely to lead to notable variations in accuracy due to the reliance on positional encoding. In contrast, for late fusion models which process each modality independently before combining them, we hypothesise minimal sensitivity to image sequencing since the fusion occurs only after individual processing. For Hybrid fusion models which integrate modalities at intermediate stages, we would expect to observe moderate sensitivity to sequencing, reflecting a partial dependence on modality order but not as extreme as early fusion models.

Additionally, dataset complexity is expected to modulate these effects. With respect to inputs, the M3Exam dataset is considerably more complex of the two benchmarks given that is contains up to five images per prompt; however, the difficulty of the actual question tends to generally be with the M3COTS dataset\footnote{ChatGPT-4 achieved an accuracy of 71.8\% on M3Exam \cite{zhang2023m3exammultilingualmultimodalmultilevel} and 62.6\% on the M3COTS dataset \cite{chen-etal-2024-m3cot} respectively using CoT in the initial experiments.}. Therefore, the increased cognitive load may reduce the model’s ability to distinguish the effects of different image positions particularly in early fusion models. On the other hand, in the M3COTS dataset, where each prompt contains only one image, we would anticipate clearer sequencing effects, as the model's attention is more focused on integrating fewer modalities. These hypotheses will be evaluated for accuracy differences across prompt configurations to assess the potential influences of both fusion strategy and dataset complexity.

\subsubsection{Evaluation}

Our experiment evaluations were mainly performed using a mix of comparing the percentage of correct responses, conducting mean rank analyses, and performing tests for statistical significance. For the statistical evaluation of binary outcomes per response (i.e. correct/incorrect), the McNemar's test was used as it is specifically designed for binary outcomes and thus provides an effective way to compare the relative performance under different conditions for the same questions.  
Mean ranks were employed to offer a more comprehensive and insightful understanding of the impact of image and text sequencing configurations. For each question type and configuration, ranks were assigned based on the accuracy performance of the LLMs, with a lower rank indicating better performance. These ranks were then averaged across different sub-categories within each dataset, such as subject domains and question types. Analysing the \textit{mean ranks} subsequently helped in identifying  more generally what the most optimal configurations tended to be by consolidating performances over all configurations. Mean ranks therefore provided another concise perspective alongside that of accuracy comparisons. 
The statistical tests provided insights but were considered merely as one of several indicators rather than the sole arbiter of significance.





\section{Results}

This section first examines the results from variations in image-text sequencing. Subsequently, it assesses the impact of the characteristics of questions on accuracy. Following this, it presents the findings from the analysis of image or instruction sequencing effects. Lastly, it considers the outcomes of the proposed priming strategy.

\subsection{Image-Text Sequence Variation}

\begin{figure}[htbp]
    \centering
    \begin{subfigure}[b]{0.48\linewidth}
        \centering
        \includegraphics[width=\linewidth]{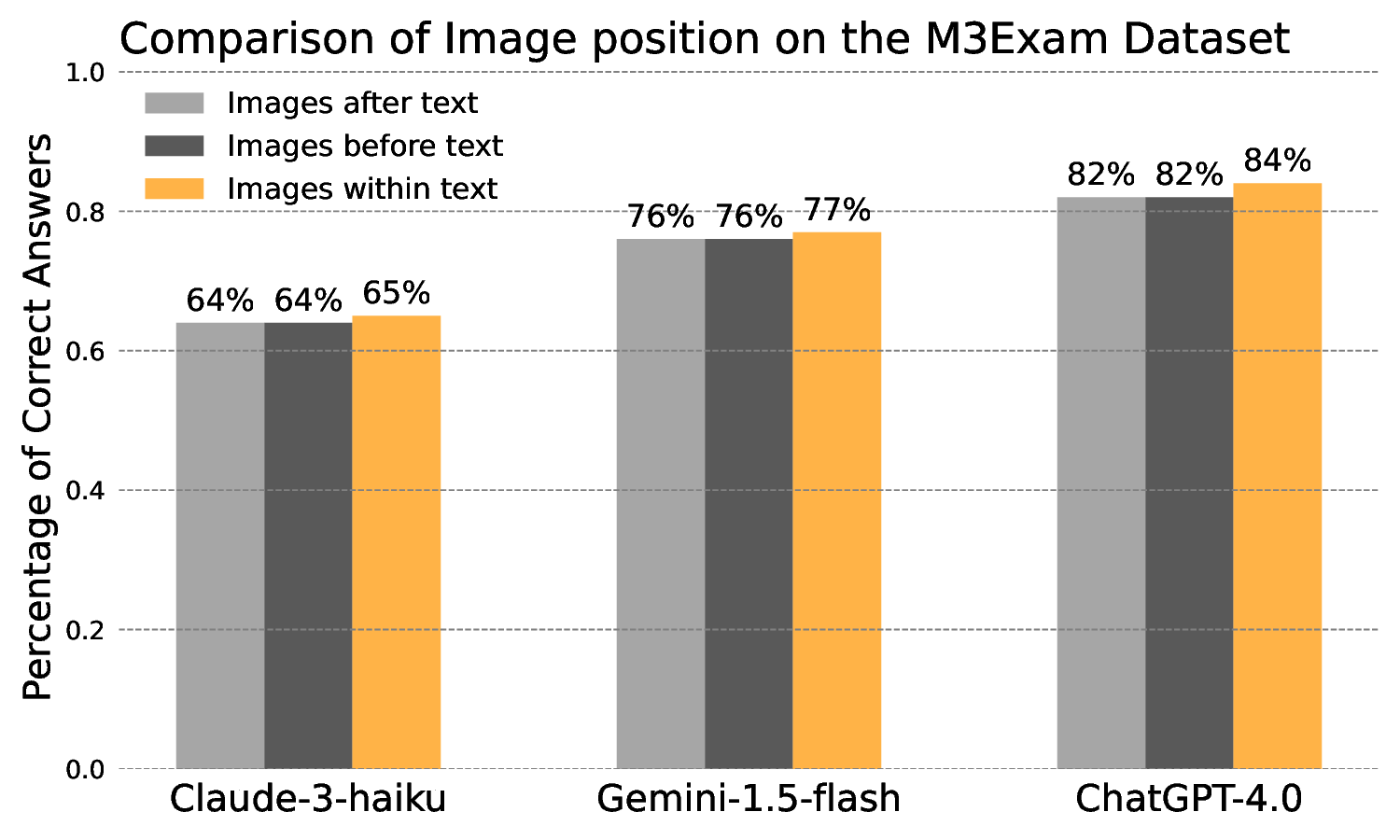}
        \caption{ M3Exam dataset}
        \label{fig:ImagePositionM3Exam}
    \end{subfigure}
    \hfill
    \begin{subfigure}[b]{0.48\linewidth}
        \centering
        \includegraphics[width=\linewidth]{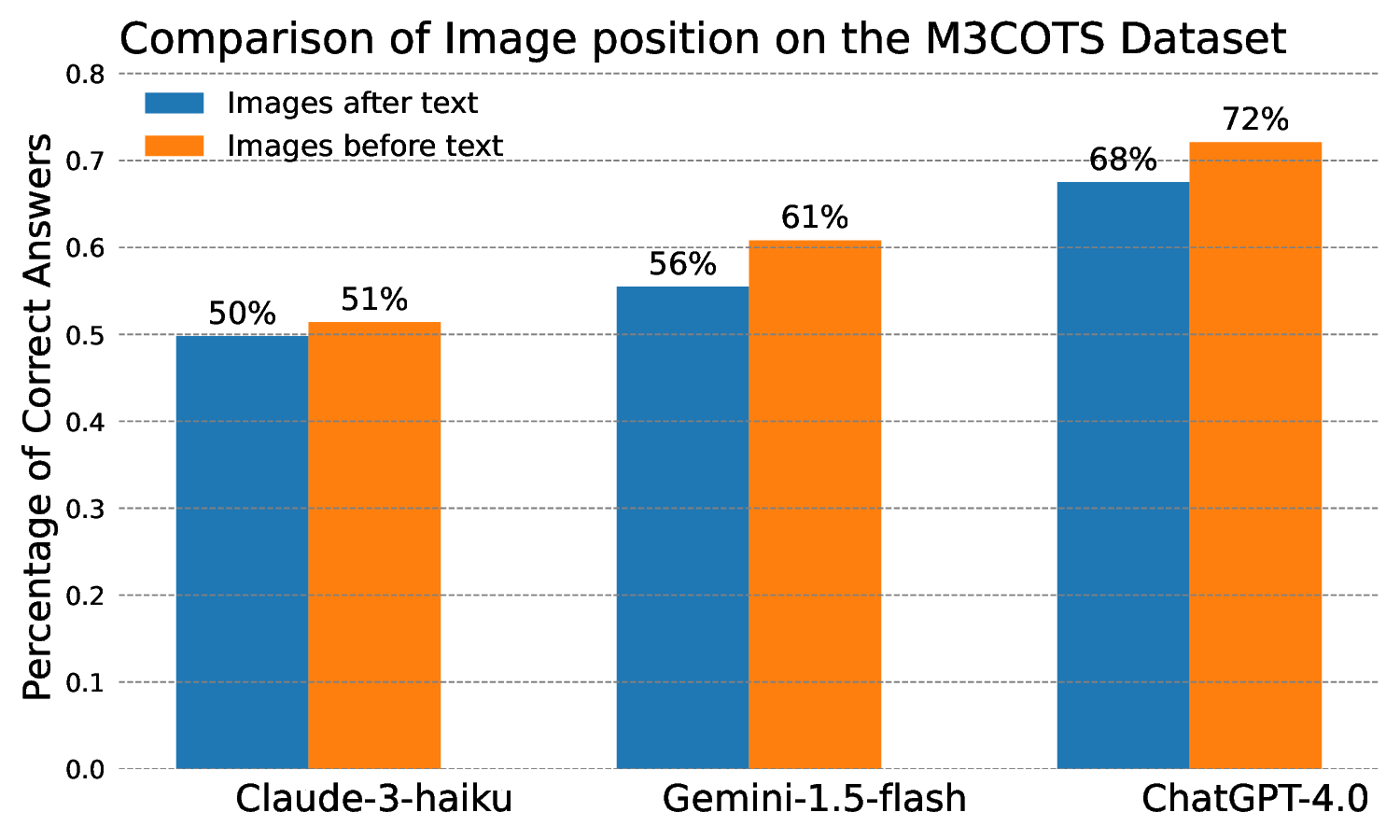}
        \caption{M3COTS dataset}
        \label{fig:ImagePositionM3COTS}
    \end{subfigure}
    \caption{Comparison of image placement positions on the M3Exam and M3COTS datasets}
    \label{fig:ImagePositionComparison}
\end{figure}

Figure \ref{fig:ImagePositionComparison} shows the accuracies of the three LLMs on both datasets, with respect to the different placements of the images in the prompt sequences. At a high level, it can be seen that generally LLMs tend to score higher on M3Exam than on M3COTS, which is in line with results in literature, which has reported  71.8\% \cite{zhang2023m3exammultilingualmultimodalmultilevel}  and 62.6\% \cite{chen-etal-2024-m3cot} respectively using the older ChatGPT-4 with CoT. ChatGPT-4o also consistently outperformed Claude-3-haiku and Gemini-1.5-flash on both datasets by a significant margin, while, Claude-3-haiku has demonstrated the lowest overall performance on both datasets. 
Across both figures, it can also be seen generally, that placing images within the text on the M3Exam dataset consistently yields higher accuracies over other placements, while on the M3COTS dataset, we see that placing the images before all the textual components (i.e. background, questions, options and other instructions) consistently improved accuracies. However, the results also show that in general the performance differences between the modality sequencing strategies were less pronounced on M3Exam than on M3COTS datasets. From this, some inferences about the possible fusion strategies can be made.

The results from Figure \ref{fig:ImagePositionComparison} across both the M3Exam and M3COTS datasets may suggest that Claude-3-haiku is likely utilising a late or hybrid fusion strategy as indicated by its stable performance (accuracies differ approx. 1\%) across different prompt configurations in both more complex (M3Exam) and simpler (M3COTS) multi-modal reasoning tasks as opposed to other models. 
The minimal sensitivity to image sequencing supports the notion that the underlying Claude-3 model processes modalities independently before merging them, leading to consistent outcomes regardless of the prompt sequencing structure. Conversely, Gemini-1.5-flash and ChatGPT-4.0 show patterns consistent with early fusion approaches. Both models exhibit greater sensitivity to prompt sequencing in the M3COTS dataset (accuracies differences range approx. 4\%-6\%), where the reasoning task is less complex given there is only one image per prompt. In contrast, the M3Exam dataset, though it has a lower degree of content-difficulty than M3COTS, given its higher input complexity comprising multiple images per prompt, this likely dampens the effects of image sequencing due to the increased cognitive load and reasoning requirements. 
This reinforces the hypothesis that early fusion models perform better when the task complexity is lower, and the modality integration can be influenced by the position of images in the prompt.


Table \ref{table:M3ExamImagePositionData} details a deeper performance profile of each sequencing configuration with respect to the different subject areas of the M3Exam dataset, and the various characteristics that questions from each discipline could influence accuracies when combined with different image placements. The summary of the table in the form of mean ranks consistently indicates that on average, placing images within the text yielded best results, while showing little difference between the \textit{before} and \textit{after} placements for all LLMs\footnote{Neither the McNemar's nor the Wilcoxon tests showed statistical significance of the results in the Table \ref{table:M3ExamImagePositionData}.}. 

\begin{table}[htbp]
\centering
\caption{Comparison of image position on the M3Exam Data}
\fontsize{10pt}{12pt}\selectfont

\label{table:M3ExamImagePositionData}
\begin{tabular}{lcccccccccccc}
\toprule

& \multicolumn{3}{c}{\textbf{Claude-3-haiku}} & \multicolumn{3}{c}{\textbf{Gemini-1.5-flash}} & \multicolumn{3}{c}{\textbf{ChatGPT-4o}} \\
\cmidrule(lr){2-4} \cmidrule(lr){5-7} \cmidrule(lr){8-10}
\textbf{Subject} & \textbf{After} & \textbf{Before} & \textbf{Within} & \textbf{After} & \textbf{Before} & \textbf{Within} & \textbf{After} & \textbf{Before} & \textbf{Within} \\
\midrule
English & 0.81 & \textbf{0.90} & \textbf{0.90} & 0.94 & \textbf{1.00} & 0.94 & 0.97 & \textbf{1.00} & \textbf{1.00} \\
Algebra1 & 0.58 & \textbf{0.42} & \textbf{0.42} & 0.42 & 0.58 & \textbf{0.63} & 0.58 & \textbf{0.68} & \textbf{0.68} \\
Algebra2 & 0.19 & \textbf{0.50} & 0.38 & 0.56 & 0.50 & \textbf{0.63} & \textbf{0.63} & 0.56 & \textbf{0.63} \\
Geometry & \textbf{0.31} & 0.29 & \textbf{0.31} & \textbf{0.49} & 0.47 & 0.47 & \textbf{0.63} & 0.59 & 0.59 \\
Math & 0.39 & 0.37 & \textbf{0.42} & 0.59 & 0.58 & \textbf{0.60} & 0.64 & \textbf{0.70} & 0.69 \\
Chemistry & 0.67 & 0.60 & \textbf{0.73} & 0.60 & \textbf{0.80} & 0.73 & \textbf{0.87} & 0.80 & 0.80 \\
Environment & 0.79 & \textbf{0.82} & 0.81 & 0.92 & 0.92 & \textbf{0.93} & \textbf{0.98} & 0.96 & 0.94 \\
Physics & \textbf{0.43} & 0.37 & 0.36 & 0.63 & 0.63 & \textbf{0.71} & 0.77 & 0.79 & \textbf{0.85} \\
Science & 0.79 & 0.79 & \textbf{0.81} & 0.88 & 0.86 & \textbf{0.89} & 0.88 & 0.89 & \textbf{0.90} \\
Earth & 0.61 & 0.61 & \textbf{0.62} & \textbf{0.70} & 0.69 & 0.68 & 0.75 & 0.73 & \textbf{0.80} \\
History & 0.94 & \textbf{0.96} & \textbf{0.96} & \textbf{1.00} & 0.98 & 0.96 & \textbf{1.00} & \textbf{1.00} & \textbf{1.00} \\
Social & \textbf{0.87} & 0.81 & 0.84 & 0.90 & 0.93 & \textbf{0.94} & 0.94 & \textbf{0.95} & \textbf{0.95} \\
\midrule
\textbf{Mean Rank} & 2.13 & 2.21 & \textbf{1.66} & 2.12 & 2.2 & \textbf{1.67} & 2.2 & 2.0 & \textbf{1.75} \\
\midrule
\bottomrule
\end{tabular}%
\end{table}

Meanwhile, a granular investigation into the effects of image placements in the M3COTS data was also performed at a subject level to complement the results from Figure \ref{fig:ImagePositionComparison} which showed that Image First approach yields the highest accuracies, resulting in \~1\%, \~5\% and \~5\% improvements for Claude-3, Gemini-1.5, and ChatGPT4o, respectively. The detailed breakdown of the results by subject 
is seen in Table \ref{table:M3COTSImagePositionData}. Since the dataset was not designed for inter-weaved sequencing of input modalities, only the image\textit{-before} and \textit{-after} configurations were explored. Across all three LLMs, the results were consistent and indicated that on average, placing images before the text yields better performances. Using the McNemar's test statistical, significance was achieved for Gemini-1.5-flash (McNemar's Test Statistic = 209.0, \(p = 0.000\)) and ChatGPT-4o (McNemar's Test Statistic = 163.0, \(p = 0.000\)), but not for the Claude-3-haiku model. 

\begin{table}[htbp]
\centering
\caption{Comparison of Image Position on the M3COTS Data}
\fontsize{10pt}{12pt}\selectfont

\label{table:M3COTSImagePositionData}
\begin{tabular}{lccccccccc}
\toprule
& \multicolumn{2}{c}{\textbf{Claude-3-haiku}} & \multicolumn{2}{c}{\textbf{Gemini-1.5-flash}} & \multicolumn{2}{c}{\textbf{ChatGPT-4o}} \\
\cmidrule(lr){2-3} \cmidrule(lr){4-5} \cmidrule(lr){6-7}
\textbf{Subject} & \textbf{After} & \textbf{Before} & \textbf{After} & \textbf{Before} & \textbf{After} & \textbf{Before}  \\
\midrule
language-science   & \textbf{0.79} & 0.73  & \textbf{0.88} & 0.84                     &\textbf{ 0.95} & 0.94\\
natural-science    & \textbf{0.53} & \textbf{0.53}  & 0.59 & \textbf{0.64}             & 0.70 &  \textbf{0.78}\\
social-science     & \textbf{0.35} & 0.32  & 0.39 & \textbf{0.45}                     & 0.55 & \textbf{0.59}\\
physical-commonsense & 0.60   & \textbf{0.82}  & 0.77 & \textbf{0.88}                & \textbf{0.86} & 0.84\\
social-commonsense   & 0.63  & \textbf{0.70} & 0.68 & \textbf{0.74}                 & 0.76 & \textbf{0.80}\\
temporal-commonsense & 0.75  & \textbf{0.80} & 0.75 & \textbf{0.87}                  & \textbf{0.89} & 0.86\\
algebra            & 0.21 & \textbf{0.31} & 0.28 & \textbf{0.35}                      & 0.44 & \textbf{0.57}\\
geometry           & 0.24 & \textbf{0.36} & 0.36 & \textbf{0.39}                      & \textbf{0.34} & 0.33\\
theory             & 0.33 & \textbf{0.38}& 0.24 & \textbf{0.43}                      & 0.29 & \textbf{0.48}\\
\midrule
\textbf{Mean Rank} & 1.72 & \textbf{1.27} & 1.88 & \textbf{1.13}   & 1.56 & \textbf{1.44} \\

\midrule
\bottomrule
\end{tabular}%
\end{table}

\subsection{Image-Text Sequence: Attribute-Based Analysis}

Here, exam question attributes were analysed for their impact on image sequencing to evaluate whether the trends observed in the overall dataset accuracies presented earlier, hold for each of the attributes. 
For the M3Exam dataset, Levels, Prompt Length and Image Types were examined, while for M3COTS Question Types, Prompt Length and Image Types were evaluated. In the case of M3Exam data, the models' performances did not show any deviations from the results in the previous section (the details of this can be seen in Appendix \ref{AppendixImageTextAttribute}). However, in the case of certain question types for the M3COTS dataset, placing the image after the text led to significantly better performance  which was contrary to the overall results in the previous section. Table \ref{table:AttributeContraryQuesTypes} shows M3COTS question types where the optimal image sequencing diverged from the results for the overall dataset. For instance, performance on the "Physics - Velocity, Acceleration, and Forces" question type showed significantly improved  with the image placed after the text for Claude-3-Haiku (McNemar’s test p-value = 0.001), similar "Grammar" showed a significance for Gemini-1.5-Flash (McNemar’s test p-value = 0.021). This finding suggest that the impact of image sequencing varies depending on the model and context and from this we can conclude that optimally matching the image sequencing for specific question types can enhance accuracies. To estimate the potential improvement in accuracy, we selected the higher accuracy value for each question type between the two sequencing configurations---either Image First (\(a_{\text{IF}} = f_{\text{IF}}(x, q)\)) or Text First (\(a_{\text{TF}} = f_{\text{TF}}(q, x)\))---in the M3COTS dataset. This method provides a theoretical upper bound on performance improvement by considering the best possible outcome for each question type, acknowledging that this picks the best results and does not necessarily correspond to a practical sequencing strategy. Based on this analysis, the overall accuracy could potentially increase by approximately \(5\%\) for Claude-3, \(3\%\) for Gemini-1.5, and \(3\%\) for ChatGPT-4o.

\begin{table}[htbp]
\centering
\caption{Question Types with Optimal Image Position Contrary to the overall Dataset}
\fontsize{8pt}{10pt}\selectfont
\label{table:AttributeContraryQuesTypes}
\resizebox{\textwidth}{!}{%
\begin{tabular}{lccccccccc}
\toprule
\textbf{Question Type} & \multicolumn{3}{c}{\textbf{Claude-3-Haiku}} & \multicolumn{3}{c}{\textbf{Gemini-1.5-Flash}} & \multicolumn{3}{c}{\textbf{GPT-4o}} \\
\cmidrule(lr){2-4} \cmidrule(lr){5-7} \cmidrule(lr){8-10}
  & \textbf{After} & \textbf{Before} & \textbf{p-value} & \textbf{After} & \textbf{Before} & \textbf{p-value} & \textbf{After} & \textbf{Before} & \textbf{p-value} \\ 

\midrule
Physics - Velocity, Acceleration, and Forces & \textbf{0.48} & 0.18 & 0.001 & 0.60 & \textbf{0.64} & 0.774 & 0.86 & \textbf{0.88} & 1.000\\ 
Geography - Climate Analysis & \textbf{0.35} & 0.19 & 0.031 & 0.38 & \textbf{0.43} & 0.754 & \textbf{0.76} & 0.59 & 0.109 \\ 
Grammar$^a$ & \textbf{0.87} & \textbf{0.87} & 1.000 & \textbf{0.96} & 0.79 & 0.021 & 0.96 & \textbf{1.00} & 0.500\\ 
\midrule
\bottomrule
\end{tabular}%
}
\textsuperscript{ $^a$ The full name of the "Grammar" question type within the datset is "Grammar-Sentences, fragments, and run-ons."}
\\
\end{table}

\begin{figure}[ht]
\caption{Example question with impacted by sequencing}
\label{fig:ExamplesQuestionsIMS}
\begin{minipage}{0.45\textwidth}
\textbf{Improved performance with Image First}
\includegraphics[width=\textwidth]{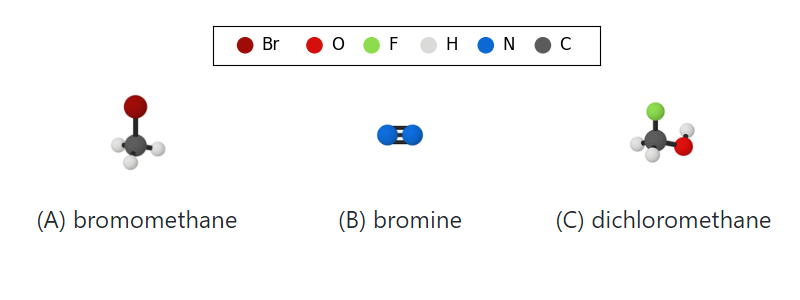}

\textbf{Question:}\\
Find the correct molecular name based on the legend.\\
Options:\\
(A) All of the answer choices are wrong.\\
(B) Option B in the image\\
(C) Option A in the image\\
(D) Option C in the image\\

In the image \textbf{(A)Bromomethane} is correct\\
Answer: (A)\textbf{\textit{ Incorrect this should be option (C)}}\\
\end{minipage}\hfill
\begin{minipage}{0.45\textwidth}
\textbf{Improved performance with Text First}
\\
\\
\textbf{Question:}\\
Which is a compound sentence?\\
Options:\\
(A) Option A in the image\\
(B) Option B in the image\\ 
(C) It seems like there's an error in all the provided options.\\
\includegraphics[width=\textwidth]{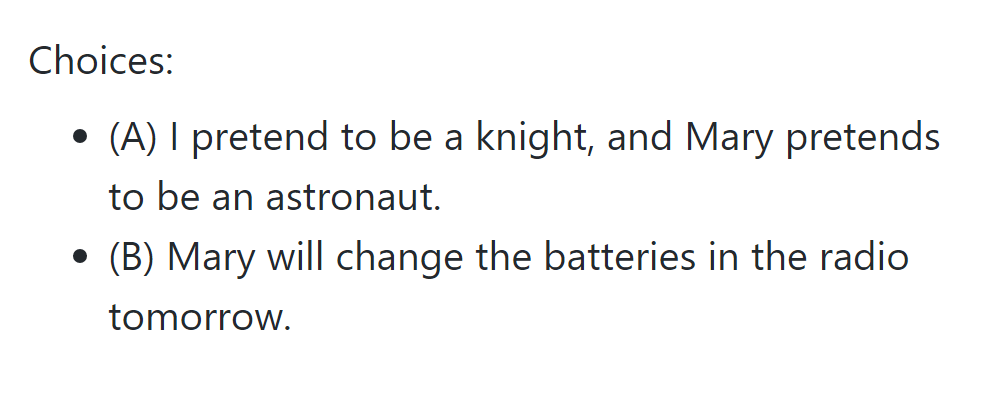}
\end{minipage}
\end{figure}

Examining the "Grammar" question type reveals a structural difference: the image presents text options for selection rather than displaying a question or additional visual content (See Figure \ref{fig:ExamplesQuestionsIMS} for an example). In contrast, question types that followed the overall dataset pattern and were most impacted by sequencing changes often involved a nested multiple-choice format, where one question referenced another. For instance, "Chemistry-Atoms and Molecules Recognise" questions (see Figure \ref{fig:ExamplesQuestionsIMS} for an example), ChatGPT's accuracy dropped from 67\% to 32\% when the image was moved from before to after the text. When the image was placed after the text the model correctly interpreted the image as it was more likely to select the option shown in the image rather than the one stated in the original text. These results suggest that the sequencing of content plays a critical role in questions where references are important.

\subsection{Image-Text Sequence: Image Versus Instructions Analysis}\label{SecImagevsInstructions}



For these experiments, we utilised a dataset comprising questions presented either solely in text or as formulas embedded within images. Specifically, we employed the "Elementary Algebra" (363 questions) and "Grammar" (205 questions) subsets from the M3COTS dataset. The primary objective was to investigate whether the sequencing of instructions—independent of image placement—affects model performance. To isolate the effect of sequencing, we extracted text from images during preprocessing, creating text-only versions of the questions. This extraction was performed using ChatGPT-4o and Gemini-1.5-flash
Table \ref{tab:ImageExtractResults} presents the performance of three multi-modal LLMs—Claude-3-Haiku, Gemini-1.5-Flash, and GPT-4o—under different sequencing conditions. In this context, "After" indicates that the image is presented after the textual instructions, while "Before" denotes that the image precedes the text. The "multi-modal" column refers to the original questions containing both images and text, whereas the "text" column represents the text-only versions.


\begin{table}[htbp]
\centering
\caption{Image-Text Sequence:Image Versus Instructions Analysis Results}
\fontsize{8pt}{10pt}\selectfont
\label{tab:ImageExtractResults}
\begin{tabular}{lccccccccc}
\toprule
\textbf{Question Type} & \multicolumn{3}{c}{\textbf{Claude-3-Haiku}} & \multicolumn{3}{c}{\textbf{Gemini-1.5-Flash}} & \multicolumn{3}{c}{\textbf{GPT-4o}} \\
\cmidrule(lr){2-4} \cmidrule(lr){5-7} \cmidrule(lr){8-10}
  & \textbf{After} & \textbf{Before} & \textbf{p-value} & \textbf{After} & \textbf{Before} & \textbf{p-value} & \textbf{After} & \textbf{Before} & \textbf{p-value} \\ 
\midrule
Elementary Algebra (multi-modal) & 0.32 & \textbf{0.39} & 0.207 & 0.27 & \textbf{0.38} & 0 & 0.45 & \textbf{0.55} & 0.001 \\ 
Elementary Algebra (text) & 0.27 & \textbf{0.35} & 0.015 & 0.25 & \textbf{0.43} & 0 & 0.35 & \textbf{0.64} & 0 \\ 
\midrule
Grammar  (multi-modal) & \textbf{0.38} & 0.35 & 0.23 & \textbf{0.94} & 0.79 & 0 & 0.95 & \textbf{0.96} & 0.508 \\ 
Grammar  (text) & \textbf{0.91} & 0.88 & 0.18 & \textbf{0.92} & 0.85 & 0.006 & 0.96 & \textbf{0.98} & 0.125 \\ 

\bottomrule
\end{tabular}%
\end{table}

The experimental results in Table \ref{tab:ImageExtractResults} reveal that the sequencing of images and text within prompts significantly influences the reasoning performance of multi-modal large language models, with effects varying by task and model. Specifically, for \enquote{Elementary Algebra} questions, both Gemini-1.5-Flash and GPT-4o demonstrated markedly higher accuracy when images were presented before textual instructions, suggesting that visual context aids mathematical reasoning. In contrast, for \enquote{Grammar} questions, Gemini-1.5-Flash achieved better performance when textual instructions preceded images, indicating that linguistic tasks benefit from a text-first approach. Claude-3-Haiku showed less sensitivity to sequencing, with only limited performance variations across different orders. Additionally, the text-only versions of the prompts mirrored these patterns, underscoring that the order of instructional information alone, independent of image placement, plays a crucial role in model performance. These findings highlight the importance of tailoring prompt structures to both the nature of the task and the specific model in use, thereby optimising multi-modal reasoning capabilities across diverse applications.

\subsection{Prompt Priming for Relationship Analysis}

The initial baseline results for M3COTS shown in Figure \ref{fig:ImagePositionM3COTS} indicated that placing an image before the textual modality yielded higher accuracies. 
To assess whether explicit priming could influence the processing order of modalities and thereby enhance model performance, we conducted prompt priming experiments across all questions in the M3COTS dataset. Specifically, we instructed the LLMs to prioritise image processing even when images were presented after textual instructions. Contrary to our hypothesis, the results in Table \ref{table:ImgaePoistionTX_IMM3COTSData} indicate that this priming strategy led to a consistent decline in accuracy across all tested models. Claude-3-Haiku's accuracy decreased from 0.51 to 0.45, Gemini-1.5-Flash from 0.56 to 0.53, and ChatGPT-4o from 0.67 to 0.64 when prompted to focus on images first despite their subsequent placement. These findings suggest that the inherent processing order of the models, likely ingrained through their training data and architectural design, is resistant to override through simple priming instructions. The decline in performance implies that the models may prioritise modalities based on their default configurations (including modality fusion strategies), making it challenging for external prompts to effectively alter their attention mechanisms.

These results underscore the role of modality sequencing over priming in prompt engineering for multi-modal LLMs. While physical ordering of information (i.e., presenting images before text) tends to enhance performance as demonstrated in our baseline experiments (Figure \ref{fig:ImagePositionM3Exam} and \ref{fig:ImagePositionM3COTS}), attempting to manipulate the processing order through priming does not yield the same benefits and may even be detrimental. This highlights a fundamental limitation in current prompt engineering techniques for the current series of LLMs, where explicit instructions alone are insufficient to change the models' inherent information processing pathways. Consequently, effective optimisation of multi-modal reasoning capabilities should prioritise the strategic sequencing of modalities within prompts. 

\begin{table}[ht]
\centering
\caption{Comparison of different Prompt Priming methods for images after text on the M3COTS Data}
\label{table:ImgaePoistionTX_IMM3COTSData}
\resizebox{\textwidth}{!}{%
\begin{tabular}{@{}lccc@{}}
\toprule
\textbf{Subject} &  Claude-3-haiku \cite{anthropic2024claude} &   Gemini-1.5-flash \cite{reid2024gemini} &   ChatGPT-4o \cite{openai2024gpt4technicalreport}\\ 
\midrule
Images after text - Baseline (Figure \ref{fig:ImagePositionM3COTS})    & \textbf{0.51} & \textbf{0.56} & \textbf{0.67} \\
Images after text - Prompt to process image first & 0.45 & 0.53 & 0.64 \\
\bottomrule
\end{tabular}
}
\end{table}

\section{Discussion}

Our research investigated the impact of varying the sequencing of images and text modalities on the reasoning performance of LLMs and found instructive results. Our work built on and extended similar investigations considering the impact of altering the relative position of words \cite{liu-etal-2024-lost,lu2021fantastically} or the instruction order in text prompts \cite{chu2024better}. We hypothesised that the order in which these modalities are sequenced would influence reasoning performance. The results confirmed this hypothesis, showing that the optimal sequencing varied depending on the dataset: placing images inline within the text yielded the best performance on the M3Exam dataset, while presenting images before the text led to superior performance on the M3COTS dataset. Further analysis showed that within the M3COTS dataset, certain question types were more sensitive to sequencing changes than others, with the optimal sequencing of modality presentation differing by question type and model. These findings suggest that both the dataset structure and the complexity of the questions influence how modality sequencing affects reasoning performance in LLMs.

\subsection{Modality sequencing and fusion strategies}

The effect of sequencing image and text modalities on LLM reasoning performance varied significantly across the two datasets, highlighting the pivotal role of instruction tuning and prompt design in shaping model behaviour , with the underlying multi-modal fusion strategies of each LLM being an unknown confounding factor (RQ1).  For the M3Exam dataset, the best performance was achieved when images were interwoven with the text. The approach mirrored the actual exam structure designed to optimise student comprehension by aligning modalities for effective information flow for humans. This same sequencing also proved beneficial for LLM reasoning for this dataset. In contrast, the M3COTS dataset, designed to challenge multi-modal reasoning, generally performed better when images were presented before the text. This suggests that placing the image first provides a visual context that aids the reasoning process, as recommended by vendors\cite{googleimageunderstanding}\cite{anthropic2024claude} \cite{OpenAI2023community}. Variations within the dataset indicate that the optimal sequencing depends on the specific structure of the individual question, highlighting that the best modality sequencing is context-dependent and shaped by both the dataset and the task at hand.

The study suggests that attention mechanisms in transformer-based LLMs likely influences modality bias which affects the reasoning performance based on the sequencing of modalities in prompts. We inferred from the results that altering the order of text and images changes attention distribution across modalities. In early fusion architectures, positional encoding causes earlier modalities to receive disproportionately higher attention, potentially underutilising later modalities and hindering effective multi-modal integration in complex reasoning tasks. These findings have practical implications for both prompt design and model development. From this insight, prompt designers may consider strategically sequencing modalities to align with the logical flow of reasoning to ensures critical information receives appropriate attention. For model developers, addressing inherent positional biases in attention mechanisms is essential as this could involve architectural adjustments or training strategies that promote equitable attention distribution across modalities.

\subsection{Question Complexity and Sequencing Sensitivity}

Analysis of the question types most impacted by sequencing changes, particularly in the M3COTS dataset, often involved a nested multiple-choice format where one question referenced another. While explored LLMs frequently succeeded in solving the underlying reasoning task related to the image, they often struggled with the final step—revisiting earlier information to select the correct option within the original question (e.g., pointing to the option list in the image). This challenge highlights issues related to multi-hop reasoning and the models' capacity to maintain context over several reasoning steps. The linear reasoning approach facilitated by CoT prompting encourages step-by-step processing but may not adequately support the backtracking required in nested questions. The transformer's positional encoding of tokens is crucial for maintaining context, but as the reasoning becomes more complex the output sequences lengthen, earlier information may receive diminished attention due to the model's attention decay over distance. In contrast, for other question types like the ones in the "Grammar" format, where the optimal image sequencing diverged from the overall dataset pattern, a structural difference was observed. Here, the image presents text options for selection, rather than displaying a question or additional visual content. The options appear within the image after the text-based question \ref{fig:ExamplesQuestionsIMS}. The flow of information matches the logical steps of reasoning, emphasising that optimal sequencing is not a one-size-fits-all approach but depends heavily on the structural and logical flow required by the task. This suggests that it is the placement and sequencing of information represented by the image, rather than the image's physical properties, that plays a crucial role in reasoning performance (RQ2).

\subsection{Information Order vs. Modality Properties}

Our experiments revealed that the sequence in which information is presented significantly influences LLM performance, outweighing the inherent properties of the modalities themselves (RQ3). By converting images to text and evaluating single-modality prompts, we found that the order—whether text precedes image or vice versa—consistently impacted accuracy, underscoring the importance of positional encoding and attention mechanisms in transformer architectures. The models' sensitivity to information sequencing varied based on their training data and fine-tuning methods, and while Claude-3-Haiku showed minimal responsiveness to sequencing changes, Gemini-1.5-Flash and GPT-4o exhibited more pronounced improvements with optimal information ordering. Additionally, our attribute-based analysis indicated that specific question types could achieve performance gains of up to 5\% by tailoring the sequencing strategy, highlighting the necessity of strategic information ordering in prompt design. These findings suggest that effective prompt engineering, aligned with both task requirements and model characteristics is essential for optimising the reasoning capabilities of multi-modal LLMs and thereby enhancing their utility across diverse applications.


\subsection{Implications and Practical Guidelines}

The findings of this study extend beyond exam-like tasks and offer valuable insights for broader AI applications. For instance, in medical image interpretation, where integrating text-based clinical notes with diagnostic images is essential, understanding how modality sequencing impacts performance could lead to improved prompt designs that enhance diagnostic accuracy in multi-modal systems. Similarly, in autonomous systems, such as self-driving cars, the ability to reason across visual inputs and textual navigational commands could be optimised by refining fusion strategies. Further, the insights from our study also provide useful guidelines for optimising prompt design in multi-modal large language models (LLMs). Key implications and practical recommendations include:

\begin{itemize}
    \item \textbf{Strategic Sequencing of Modalities:}
    \begin{itemize}
        \item \textbf{Align with Task Requirements:} Tailor the order of images and text based on the nature and complexity of the task. For tasks requiring visual or spatial reasoning, presenting images first can provide the necessary context, whereas embedding images within text may enhance tasks that depend on contextual integration.
    \end{itemize}
    
    \item \textbf{Prioritise Physical Order Over Priming:}
    \begin{itemize}
        \item \textbf{Effective Prompt Engineering:} The physical sequencing of information has a more significant impact on model performance than relying solely on priming instructions. Ensuring that critical information is presented in an optimal order enhances attention distribution and information encoding within transformer architectures.
    \end{itemize}
    
    \item \textbf{Model-Specific Prompt Design:}
    \begin{itemize}
        \item \textbf{Adapt to Model Sensitivities:} Different models may respond uniquely to sequencing based on their training data and fine-tuning processes. Prompt designers should develop model-specific strategies to maximise reasoning accuracy by understanding each model's inherent processing tendencies.
    \end{itemize}
    
    \item \textbf{Enhance Multi-Hop Reasoning:}
    \begin{itemize}
        \item \textbf{Maintain Contextual Flow:} Proper sequencing can improve context retention and reduce attention decay, especially in multi-hop reasoning tasks. Aligning the information order with the logical steps required for reasoning helps models maintain coherence across multiple reasoning steps.
    \end{itemize}
    
    \item \textbf{Optimise Information Encoding:}
    \begin{itemize}
        \item \textbf{Consider Positional Encoding:} Recognise that the arrangement of information blocks influences how data is weighted and integrated. Strategically positioning modalities to match the logical flow of tasks can lead to significant performance improvements.
    \end{itemize}
    
    \item \textbf{Address Fusion Strategy and Modality Bias:}
    \begin{itemize}
        \item \textbf{Mitigate Positional Biases:} In early fusion architectures, positional encoding can cause earlier modalities to receive disproportionately higher attention, potentially underutilising later modalities. Model developers could consider architectural adjustments or training strategies that promote equitable attention distribution across modalities to enhance multi-modal integration.
        \item \textbf{Strategic Modality Alignment:} Prompt designers should align the sequencing of modalities with the logical flow of reasoning to ensure critical information receives appropriate attention. This alignment may mitigate the adverse effects of inherent positional biases in attention mechanisms.
    \end{itemize}
\end{itemize}



\subsection{Limitations}
The results in this study are based on the specific datasets used, namely M3COTS and M3Exam. It is important to note that datasets can introduce biases, and measuring the level of reasoning can be challenging \cite{mcintosh2024inadequacieslargelanguagemodel}. The effectiveness of image and text sequencing may differ with other datasets or question types, potentially limiting the scope of our findings. Additionally, the study focused exclusively on English-language datasets, and the behaviour of multilingual datasets remains unexplored.
The study does not provide a definitive method for identifying the ideal sequencing arrangement for a given set of instructions within a prompt; however, it demonstrates that these are factors which influence performance which require further investigation.


\subsection{Future Research}
Future investigations should explore how positional encoding interacts with reasoning steps, aiming to refine encoding techniques to better align with the logical steps required for complex reasoning tasks. Such research could provide insights into optimising positional encoding strategies to enhance LLM performance in both text and multi-modal reasoning scenarios. The effectiveness of image and text sequencing may vary with different datasets, question types, or across different languages. Given that only a limited number of question types were significantly affected by the sequencing of modalities, future studies should aim to identify which question types or structures are influenced by modality sequencing. This could involve analysing a broader range of datasets, across multiple languages and question types or various exam question formats to uncover specific patterns related to image positioning and testing these patterns across more diverse datasets.
Future research should also consider whether changing the modalities for example converting the text portion of the prompt to an image could impact reasoning performance. Alternatively, breaking the modalities up to find the optimal sequencing in the input could be another avenue for improving reasoning. Exploring these ideas could provide further insights into optimising multi-modal reasoning tasks.


\section{Conclusion}

This study explored how the sequencing of images and text in multi-modal prompts affects the reasoning performance of LLMs, particularly in exam-like tasks but with a broad applicability to other domains. Our findings indicate that the impact of modality sequencing is context-dependent, with task complexity playing a significant role. For simpler tasks with single image-questions, we observed that sequencing had a noticeable effect on performance. However, for more complex tasks that involved numerous image inputs, the high reasoning demands appeared to reduce the impact of modality ordering within the prompts. 
The study also highlighted that the question structure, particularly nested and multi-step questions, strongly influenced the effect of modality sequencing on LLM performance. While models excelled in the early steps of reasoning, they struggled when required to revisit previous information, reflecting challenges related to multi-hop reasoning and memory limitations within transformer architectures. This suggests that the logical flow of information, more than the position of the modalities themselves, can influence outcomes. Our research emphasised the importance of designing multi-modal prompts that align with the logical reasoning steps of a given task together with other recommendations. The insights from this work contribute to the development of more effective multi-modal systems, with implications for various fields that require sophisticated cross-modal reasoning.




\bibliographystyle{unsrtnat}
\bibliography{references}  

\newpage
\appendix

\section{Appendix:Image-Text Sequence:Attribute-Based Analysis}
\label{AppendixImageTextAttribute}

The following attributes were analysed for their impact on image sequencing, to evaluate whether the trends observed in the overall dataset hold for each of the attributes. For M3Exam: Levels, Prompt Length and Image types. For M3COTS Question types, Prompt Length and Image types.

\subsection*{M3Exam Data}

The models’ performance were consistent with the results for the overall dataset across the different subgroup attributes.Where contrary results were observed, they were not found to be significant according to McNemar’s test.This included images types, levels and varying input prompt lengths across all three model.  See Table \ref{table:ExamImageType} for details of the Image Type results and Table \ref{table:ExamLevels} for details of the levels.

\begin{table}[htbp]
\centering
\caption{Comparison of Image Types on the M3Exam Data Using McNemar's Test}
\label{table:ExamImageType}
\resizebox{\textwidth}{!}{%
\begin{tabular}{@{}lccccccc@{}}
        \toprule
        \textbf{} & \textbf{Images After Text} & \textbf{Images Before Text} & \textbf{Images Within Text} & \textbf{Images After vs Images Before} & \textbf{Images After vs Inline} & \textbf{Images Before vs Inline} \\ 
        \midrule
        \rowcolor{lightgray} \multicolumn{7}{c}{ChatGPT-4o \cite{openai2024gpt4technicalreport}} \\   
        Images Only & 0.80 & 0.78 & 0.80 & (Stat. = 20.0, \(p = 0.652\)) & (Stat. = 13.0, \(p = 0.585\)) & (Stat. = 19.0, \(p = 1.000\)) \\ 
        Mixture & 0.84 & 0.83 & 0.86 & (Stat. = 23.0, \(p = 0.775\)) & (Stat. = 21.0, \(p = 0.169\)) & (Stat. = 16.0, \(p = 0.054\)) \\ 
        Text Only & 1.00 & 1.00 & 1.00 & (Stat. = 0.0, \(p = 1.000\)) & (Stat. = 0.0, \(p = 1.000\)) & (Stat. = 0.0, \(p = 1.000\)) \\	
        \midrule
        \rowcolor{lightgray} \multicolumn{7}{c}{Gemini-1.5-Flash \cite{reid2024gemini}} \\
        Images Only & 0.69 & 0.71 & 0.72 & (Stat. = 22.0, \(p = 0.665\)) & (Stat. = 19.0, \(p = 0.542\)) & (Stat. = 19.0, \(p = 0.875\)) \\ 
        Mixture & 0.78 & 0.78 & 0.80 & (Stat. = 43.0, \(p = 1.000\)) & (Stat. = 32.0, \(p = 0.349\)) & (Stat. = 27.0, \(p = 0.314\)) \\ 
        Text Only & 1.00 & 1.00 & 1.00 & (Stat. = 0.0, \(p = 1.000\)) & (Stat.= 0.0, \(p = 1.000\)) & (Stat. = 0.0, \(p = 1.000\)) \\	
        \midrule
        \rowcolor{lightgray} \multicolumn{7}{c}{Claude-3-Haiku \cite{anthropic2024claude}} \\
        Images Only & 0.53 & 0.56 & 0.55 & (Stat. = 19.0, \(p = 0.193\)) & (Stat. = 23.0, \(p = 0.576\)) & (Stat. = 20.0, \(p = 0.551\)) \\ 
        Mixture & 0.70 & 0.67 & 0.70 & (Stat. = 38.0, \(p = 0.170\)) & (Stat. = 41.0, \(p = 1.000\)) & (Stat. = 29.0, \(p = 0.125\)) \\ 
        Text Only & 1.00 & 1.00 & 1.00 & (Stat. = 0.0, \(p = 1.000\)) & (Stat. = 0.0, \(p = 1.000\)) & (Stat.= 0.0, \(p = 1.000\)) \\	
        \bottomrule        
\end{tabular}
}
\end{table}

\begin{table}[htbp]
\centering
\caption{Comparison of Levels on the M3Exam Data Using McNemar's Test}
\label{table:ExamLevels}
\resizebox{\textwidth}{!}{%
\begin{tabular}{@{}lccccccc@{}}
        \toprule
        \textbf{} & \textbf{Images After Text} & \textbf{Images Before Text} & \textbf{Images Within Text} & \textbf{Images After vs Images Before} & \textbf{Images After vs Inline} & \textbf{Images Before vs Inline} \\ 
        \midrule
        \rowcolor{lightgray} \multicolumn{7}{c}{ChatGPT-4o \cite{openai2024gpt4technicalreport}} \\   
        High School (USA) & 0.80 & 0.79 & 0.82 & (Stat. = 31.0, \(p = 0.470\)) & (Stat. = 23.0, \(p = 0.092\)) & (Stat. = 25.0, \(p = 10.427\)) \\ 
        Middle School (USA) & 0.84 & 0.86 & 0.85 & (Stat. = 5.0, \(p = 0.424\)) & (Stat. = 7.0, \(p = 1.000\)) & (Stat. = 6.0, \(p = 0.607\)) \\ 
        Elementary School (USA) & 0.85 & 0.90 & 0.86 & (Stat. = 3.0, \(p = 0.344\)) & (Stat. = 3.0, \(p = 1.000\)) & (Stat. = 3.0, \(p = 0.227\)) \\	
        \midrule
        \rowcolor{lightgray} \multicolumn{7}{c}{Gemini-1.5-Flash \cite{reid2024gemini}} \\
        High School (USA)  & 0.72 & 0.72 & 0.73 & (Stat. = 45.0, \(p = 0.917\)) & (Stat. = 35.0, \(p = 0.567\)) & (Stat. = 36.0, \(p = 0.500\)) \\ 
        Middle School (USA)  & 0.80 & 0.80 & 0.82 & (Stat. = 10.0, \(p = 1.000\)) & (Stat. = 6.0, \(p = 0.454\)) & (Stat. = 7.0, \(p = 0.481\)) \\ 
        Elementary School (USA)  & 0.84 & 0.86 & 0.87 & (Stat. = 10.0, \(p = 0.832\)) & (Stat.= 5.0, \(p = 1.000\)) & (Stat. = 8.0, \(p = 0.648\)) \\	
        \midrule
        \rowcolor{lightgray} \multicolumn{7}{c}{Claude-3-Haiku \cite{anthropic2024claude}} \\
        High School (USA)  & 0.60 & 0.60 & 0.60 & (Stat. = 45.0, \(p = 1.000\)) & (Stat. = 43.0, \(p = 1.000\)) & (Stat. = 44.0, \(p = 1.000\)) \\  
        Middle School (USA)  & 0.72 & 0.72 & 0.74 & (Stat. = 14.0, \(p = 1.000\)) & (Stat. = 9.0, \(p = 0.523\)) & (Stat. = 12.0, \(p = 0.572\)) \\ 
        Elementary School (USA)  & 0.69 & 0.64 & 0.69 & (Stat. = 7.0, \(p = 0.359\)) & (Stat. = 2.0, \(p = 0.180\)) & (Stat.= 8.0, \(p = 1.000\)) \\	
        \bottomrule        
\end{tabular}
}
\end{table}

\subsection*{M3COTS Data}

For M3COTS image types and prompt lengths, performance remained consistent with the overall dataset across all three models. Where contrary results were observed, they were not significant according to McNemar’s test. See Tables \ref{table:ImgaeTypeM3CotData} for details of the Image Type results.

\begin{table}[htbp]
\centering
\caption{Comparison of Image Types on the M3Cots Data using the McNemar's test}
\fontsize{8pt}{10pt}\selectfont
\label{table:ImgaeTypeM3CotData}
\begin{tabular}{@{}lccccc@{}}
        \toprule
        \textbf{} & \textbf{Images after text} & \textbf{Images before text} & \textbf{No of Questions} & \textbf{Statistic} & \textbf{P-value} \\ 
        \midrule
        \rowcolor{lightgray} \multicolumn{6}{c}{ChatGPT-4o \cite{openai2024gpt4technicalreport}} \\
        Images Only & 0.58 & 0.64 & 239 & 12.00 & 0.034 \\ 
        Mixture & 0.65 & 0.70 & 1507 & 121.0 & 0.000 \\ 
        Text Only & 0.81 & 0.85 & 378 &  30.00 & 00.009 \\  
        \midrule
        \rowcolor{lightgray} \multicolumn{6}{c}{Gemini-1.5-flash \cite{reid2024gemini}} \\
        Images Only & 0.51 & 0.54 & 239 & 20.00 & 0.253 \\ 
        Mixture & 0.53 & 0.58 & 1507 & 149.0 & 0.000 \\ 
        Text Only & 0.68 & 0.72 & 378 & 40.00 & 0.002 \\
        \midrule
        \rowcolor{lightgray} \multicolumn{6}{c}{Claude-3-haiku \cite{anthropic2024claude}} \\
        Images Only  & 0.43 & 0.45 & 239 & 26.00 & 0.597 \\ 
         Mixture  & 0.47 & 0.48 & 1507 & 194.0 & 0.486 \\ 
        Text Only  & 0.62 & 0.65 & 378 & 49.00 & 0.135 \\ 
        \bottomrule        
\end{tabular}
\end{table}
\subsection*{Comparison of Prompt Lengths}

The tables present a comparison of total input prompt lengths against the accuracy of the answers. The prompt lengths include the token counts for both the image and text components as indicated by each model. For the Gemini model, the image length has a standard token length of 258 tokens.

\begin{figure}[ht]
\caption{Comparison of Prompt Lengths Across Various Sequencing Configurations for ChatGPT-4o}
\label{fig:PromptLenChatGPT}
\begin{minipage}{0.4\textwidth}
\includegraphics[width=\textwidth]{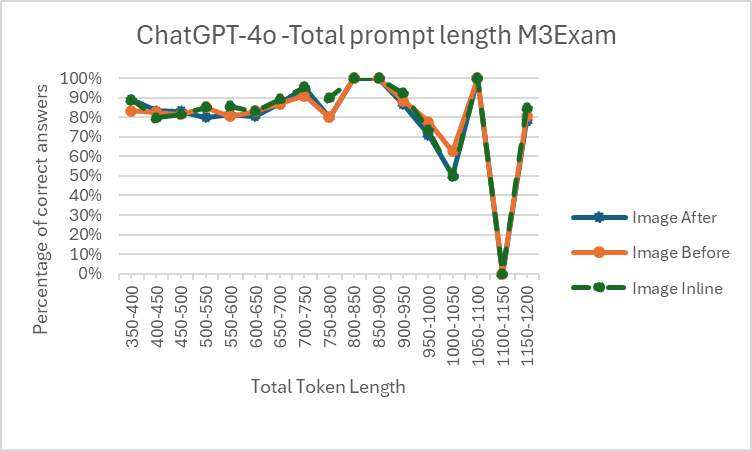}
\end{minipage}\hfill
\begin{minipage}{0.4\textwidth}
\includegraphics[width=\textwidth]{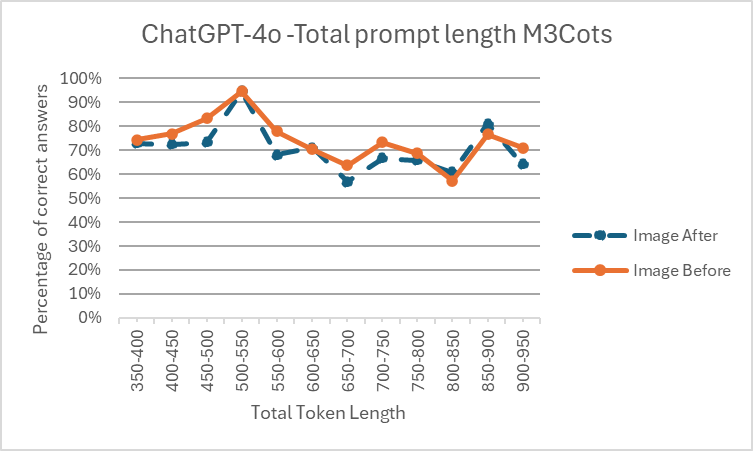}
\end{minipage}
\end{figure}

\begin{figure}[ht]
\caption{Comparison of Prompt Lengths Across Various Sequencing Configurations for Gemini-1.5-flash}
\label{fig:PromptLenGemini}
\begin{minipage}{0.4\textwidth}
\includegraphics[width=\textwidth]{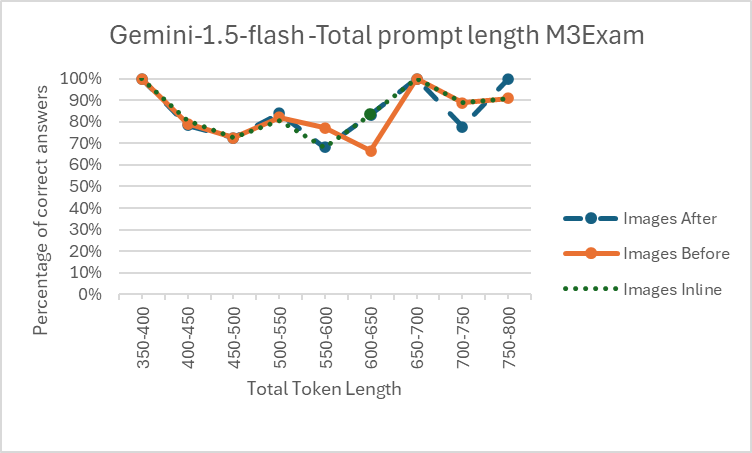}
\end{minipage}\hfill
\begin{minipage}{0.4\textwidth}
\includegraphics[width=\textwidth]{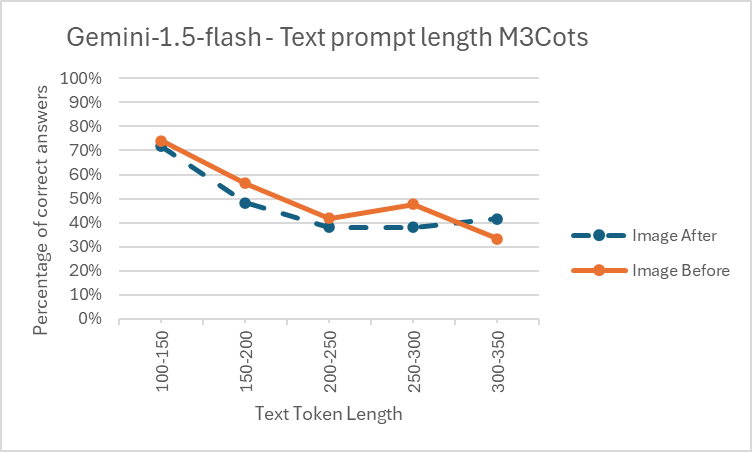}
\end{minipage}
\end{figure}

\begin{figure}[ht]
\caption{Comparison of Prompt Lengths Across Various Sequencing Configurations for Claude-3-haiku}
\label{fig:PromptLenClaude}
\begin{minipage}{0.4\textwidth}
\includegraphics[width=\textwidth]{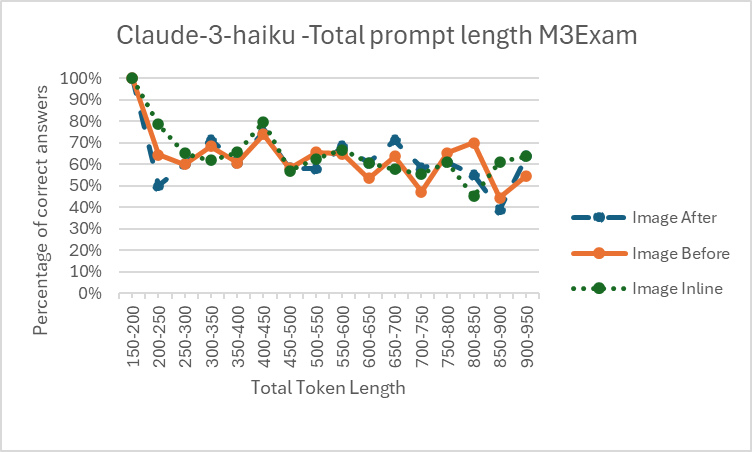}
\end{minipage}\hfill
\begin{minipage}{0.4\textwidth}
\includegraphics[width=\textwidth]{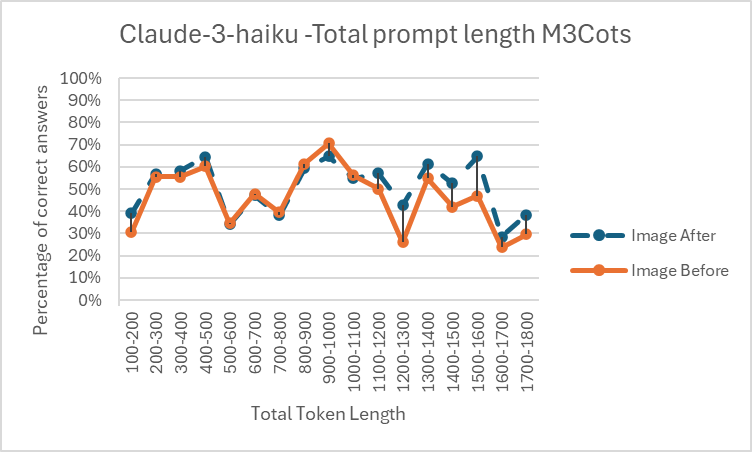}
\end{minipage}
\end{figure}

\clearpage
\end{document}